%% file: ms.tex
\documentclass[journal]{IEEEtran}

\usepackage{cite}
\usepackage{hyperref}

\ifCLASSINFOpdf
   \usepackage[pdftex]{graphicx}
\else
\fi
\usepackage{amsmath}
\usepackage{pifont}
\usepackage{amssymb}
\usepackage[table]{xcolor} 
\definecolor{electricpink}{rgb}{0.996, 0.078, 0.574}
\definecolor{darkblue}{rgb}{0, 0, 0.542}
\definecolor{bluette}{rgb}{0, 0.4470, 0.7410}
\usepackage{multirow, makecell}
\usepackage{multicol}
\usepackage{mathtools}
\usepackage{hhline}
\usepackage{tcolorbox}
\newtcbox{\inlinebox}[1][]{
 box align=base,
 nobeforeafter,
 colback=white,
 colframe=black,
 size=small,
 left=0pt,
 right=0pt,
 boxsep=2pt,
 #1}
\usepackage{booktabs}
\newcolumntype{P}[1]{>{\centering\arraybackslash}p{#1}}
\usepackage{tikz}
\usepackage[ruled]{algorithm}
\usepackage{algpseudocode}
\usepackage{subcaption}
\usepackage{balance}	
\usepackage{soul} 
\usepackage[font={small}]{caption}

\usepackage{tabularx}

\begin{document}
%
\title{A Unified Architecture for Dynamic Role Allocation and Collaborative Task Planning in Mixed Human-Robot Teams}
%
%
%


\author{Edoardo~Lamon,~\IEEEmembership{Member,~IEEE,}     Fabio~Fusaro\IEEEmembership{}, Elena~De~Momi,~\IEEEmembership{Member,~IEEE,} and~Arash~Ajoudani,~\IEEEmembership{Member,~IEEE,}%
\thanks{E. Lamon, F. Fusaro, and A. Ajoudani are with the Human-Robot Interfaces and Physical Interaction (HRI$^2$), Istituto Italiano di Tecnologia, Via  San Quirico 19d, 16163, Genova, Italy. {\tt\small edoardo.lamon@iit.it}}
\thanks{E. Lamon is also with the Department of Information Engineering and Computer Science, Universit\`a di Trento, Trento, Italy.}
\thanks{F. Fusaro and E. De Momi are with the Department of Electronics, Information and Bioengineering, Politecnico di Milano, Milano, Italy.} 
}

\maketitle

\begin{abstract}
The growing deployment of human-robot collaborative processes in several industrial applications, such as handling, welding, and assembly, unfolds the pursuit of systems that are able to manage large heterogeneous teams and, at the same time, monitor the execution of complex tasks.
In this paper, we present a novel architecture for dynamic role allocation and collaborative task planning in a mixed human-robot team of arbitrary size. The architecture capitalizes on a centralized reactive and modular task-agnostic planning method based on Behavior Trees (BTs), in charge of action scheduling, while the allocation problem is formulated through a Mixed-Integer Linear Program (MILP), which dynamically assigns individual roles or collaborations to the agents of the team. Different metrics used as MILP cost allow the architecture to favor various aspects of the collaboration (e.g. makespan, ergonomics, human preferences).
Human preferences are identified through a negotiation phase, in which a human agent can accept or refuse to execute the assigned task.
In addition, bilateral communication between humans and the system is achieved through an Augmented Reality (AR) custom user interface that provides intuitive functionalities to assist and coordinate workers in different action phases.
The computational complexity of the proposed methodology outperforms literature approaches in industrial-sized jobs and teams (problems up to 50 actions and 20 agents in the team with collaborations are solved within 1\;s). The different roles allocated as the cost functions change, highlight the flexibility of the architecture to meet several production requirements. Finally, the subjective evaluation demonstrates the high usability level and suitability for the targeted scenario.

\end{abstract}

\begin{IEEEkeywords}
Human-Robot Teaming, Task Planning, Human-Robot Collaboration. 
\end{IEEEkeywords}

%
\IEEEpeerreviewmaketitle

%
%
%
%


\section{Introduction}
\input{sections/introduction}

\section{Methodology}
\input{sections/methodology}

\section{Experiments}
\input{sections/experiments}


\section{Discussion and Conclusion}
\input{sections/discussion}

\section*{Acknowledgment}
This work was supported by the European Research Council's (ERC) starting grant Ergo-Lean (GA 850932). We acknowledge also the support of the MUR PNRR project FAIR - Future AI Research (PE00000013) funded by the NextGenerationEU.

\appendices
\input{sections/appendix}

\ifCLASSOPTIONcaptionsoff
  \newpage
\fi

\balance
\bibliographystyle{IEEEtran}
\bibliography{biblio}

\end{document}

%% file: sections/introduction.tex

\IEEEPARstart{M}{ulti-Agent} Task Allocation (MATA), representing a more general class of the well-known Multi-Robot Task Allocation~\cite{campbell2011multi}, is the problem of determining which agent, human or robot, while complying with task constraints and agent features, is in charge (and when) to execute every single action required to achieve the team's goal. This issue is growing in significance, especially in the field of Human-Robot Collaboration (HRC), with the aim of improving performance by creating synergistic collaborations between human and robotic agents~\cite{ajoudani2018progress,villani2018survey}.
Role allocation methods have applications in different fields, such as environmental exploration and monitoring~\cite{farinelli2004multirobot}, rescue~\cite{kitano2001robocup}, surveillance, transportation~\cite{tambe1997towards}, and robotic soccer~\cite{tambe1997towards, stone1999task, farinelli2004multirobot}. 


Challenging 
role allocation problems are characterized mainly by the assortment of tasks\footnote{In this paper, the term \textit{task} and \textit{role} are used as synonyms, as prescribed by \textit{Gerkey and Matari\'c}, since \textit{"the underlying problem remains the same"}~\cite{gerkey2003role}.} and large team sizes. MATA problems further require suitable mechanisms to capture the heterogeneity of the agents~\cite{notomista2022resilient}. 
MATA problems further require suitable mechanisms to capture the heterogeneity of the agents~\cite{notomista2022resilient} and the advantages of collaborations between agents. Moreover, industrial tasks (e.g., machine tending, materials handling, assembly, etc.) are generally characterized by several schedule constraints in terms of time, space, and resources. While often task allocation and task planning are treated as separate problems, they are actually two essential aspects of the same domain.
A suitable architecture that faces these problems should be able to model a wide variety of tasks while not requiring prior knowledge on the plan domain (\textit{task agnostic}), break down a complex task into its atomic actions, and optimally assign them on the fly (\textit{dynamic allocation}), to a single agent or to a group of agents (\textit{collaborative roles}), according to \textit{multiple optimization criteria}. This should be obtained while complying with online contingencies and plan variations caused by human presence and potential failures (\textit{variable sequence}). For this reason, \textit{scalability} for large tasks and teams, and \textit{re-planning capabilities} are paramount.

In light of this, in this work we propose a novel unified MATA architecture to execute complex tasks in a mixed human-robot team of arbitrary size. The architecture capitalizes on a centralized reactive and modular task-agnostic planning method based on Behavior Trees (BTs), in charge of scheduling the actions, and on a MILP-based role allocator that dynamically assigns individual and collaborative roles to the agents of the team.
To integrate human preferences into the allocation loop, a negotiation phase is introduced before action execution, in which each human worker can accept or refuse to execute the assigned task. 
Bilateral communication between humans and the system is achieved through an Augmented Reality (AR) custom user interface that provides intuitive functionalities to assist and coordinate workers in different action execution phases. The interface receives instructions and visual information on the assigned task and allows workers to give feedback to the system about their decisions.

\subsection{Related Works}
\input{sections/related_works}

\subsection{Contribution}
\input{sections/contribution}

%% file: sections/related_works.tex
\begin{table}[!t]
\setlength\extrarowheight{3.5pt}
\begin{center}
\resizebox{\columnwidth}{!}{\begin{tabular}{c|ccccccc}
Paper\textbackslash Feature & \rotatebox[origin=c]{90}{Task} \rotatebox[origin=c]{90}{Agnostic} & \rotatebox[origin=c]{90}{Variable} \rotatebox[origin=c]{90}{Sequence} & \rotatebox[origin=c]{90}{Dynamic} \rotatebox[origin=c]{90}{Allocation} & \rotatebox[origin=c]{90}{Re-planning} \rotatebox[origin=c]{90}{Capability} & \rotatebox[origin=c]{90}{Agents} \rotatebox[origin=c]{90}{Scalability} & \rotatebox[origin=c]{90}{Collaborative} \rotatebox[origin=c]{90}{Roles} & \rotatebox[origin=c]{90}{Multiple Opt.} \rotatebox[origin=c]{90}{Criteria} \\[0.7mm] 
\hline
Chen \textit{et al.}, \cite{chen2014optimal} & \ding{55} & \ding{51} & \ding{51} & \ding{55} & \makecell{only\\robots} & \ding{55} & \ding{55}\\[0.7mm] 
Johannsmeier \textit{et al.} \cite{johannsmeier2017hierarchical} & \ding{55} & \ding{55} & \ding{55} & \ding{55} & \ding{51} & \ding{55} & \ding{51}\\[0.7mm] 
Nikolakis \textit{et al.} \cite{nikolakis2018dynamic} & \ding{51} & \ding{51} & \ding{55} & \ding{51} & N/D & \ding{51} & \ding{51}\\[0.7mm] 
Bruno \textit{et al.} \cite{bruno2018dynamic} & \ding{51} & \ding{55} & \ding{51} & \ding{55} & \ding{55} & \ding{51} & \ding{51}\\[0.7mm] 
Michalos \textit{et al.} \cite{michalos2018method} & \ding{55} & \ding{55} & \ding{55} & \ding{55} & \ding{55} & \ding{55} & \ding{51}\\[0.7mm] 
Gombolay \textit{et al.} \cite{gombolay2018fast} & \ding{51} & \ding{51} & \ding{55} & \ding{51} & \ding{51} & \ding{55} & \ding{51}\\[0.7mm] 
Casalino \textit{et al.} \cite{casalino2019optimal} & \ding{55} & \ding{51} & \ding{55} & \ding{51} & N/D & \makecell{only\\preset} & \ding{55}\\[0.7mm] 
El Makrini \textit{et al.} \cite{el2019task} & \ding{55} & \ding{51} & \ding{51} & \ding{55} & \ding{55} & \ding{55} & \ding{51}\\[0.7mm] 
Pupa \textit{et al.} \cite{pupa2021human} & \ding{51} & \ding{51} & \ding{55} & \ding{51} & \ding{55} & \ding{55} & \ding{51}\\[0.7mm] 
Darvish \textit{et al.} \cite{darvish2021hierarchical} & \ding{55} & \ding{55} & \ding{51} & \ding{51} & \ding{51} & \ding{55} & \ding{51}\\[0.7mm] 
Lee \textit{et al.} \cite{lee2022task} & \ding{55} & \ding{51} & \ding{55} & \ding{55} & \ding{55} & \ding{51} & \ding{51}\\[0.7mm] 
\textbf{Ours} & \ding{51} & \ding{51} & \ding{51} & \ding{51} & \ding{51} & \ding{51} & \ding{51}\\[0.7mm] 
\hline
\end{tabular}}
\end{center}
\vspace{-2mm}
\caption{Comparison with relevant prior works.}
\label{table:literature_comparison}
\vspace{-5mm}
\end{table}

The problem of allocating roles in a homogeneous robotic team has been thoroughly investigated in the literature. 
Combinatorial optimization methods, such as Constraint Programming and Mixed-Integer Linear Programming (MILP)~\cite{gerkey2004formal, korsah2013comprehensive}, can scale up problems of different sizes easily and account for different schedule dependencies, such as temporal constraints and limited resources. However, a solution to the MATA problem for large-scale industrial tasks in mixed human-robot teams remains NP-hard with exponential complexity~\cite{korsah2013comprehensive}.
Examples of such formulation, which take inspiration from real-time processor scheduling techniques~\cite{gombolay2018fast, ferreira2021scheduling, lee2022task}, aim to minimize the overall task completion time. 
Nevertheless, in human-populated environments, optimally assigning tasks using only task duration as a metric is not sufficient, and often it is necessary to deviate from the original plan. For these reasons, other indicators have been proposed to dynamically modify the plan. For instance, human preferences can be included and updated during task execution~\cite{gombolay2015coordination}.
In \cite{rahman2018mutual}, a two-level feedforward optimization is introduced. In the first offline phase, all the possible feasible allocations are computed. Then, in the online phase, the task is rescheduled according to the trust measurement between the agents.
Similarly, \textit{Li et al.}~\cite{li2019sequence} solve the role allocation problem of a disassembly task through a genetic metaheuristic algorithm that searches for the optimal or near-optimal solution in the feasible solution space by minimizing the total disassembly time. Again, an online strategy is employed to switch team members according to human fatigue. In both cases, the experimental results highlight the importance of allocation algorithms to be able to re-plan according to online measured features during task execution.

However, the computation time of such algorithms remains reasonably low only for small-sized problems. To speed up the search and simplify the problem size, researchers began to employ different task models that could decompose the task into subtasks of smaller size.
In~\cite{pupa2021human} the task is represented by a directed acyclic graph. The graph models the dependency relationship between subtasks by connecting subtasks in a layered structure. Some of them could be independent of each other, so no path goes from one subtask to the next. The graph is then rearranged so that all parallel tasks are grouped into several sets (levels). The allocation is solved through a MILP that considers the levels in his formulation.
Another extensively used method to model industrial assembly (and disassembly) tasks is represented by AND/OR graphs. They allow a compact representation of complex assembly tasks and a complete graph search to generate feasible assembly sequences. 
Recently, AND/OR graphs were extended to the multi-agent scenario in human-robot collaborative processes~\cite{hawkins2014anticipating, johannsmeier2017hierarchical, darvish2018flexible, merlo2022dynamic}. For instance, by including different costs in the AND/OR graph hyperarcs, it is possible to represent the execution by different agents, including also agents characteristics~\cite{lamon2019capability}. The optimal assembly/agent sequence can be obtained offline~\cite{johannsmeier2017hierarchical} or online~\cite{merlo2022dynamic} through best-first search algorithms such as AO*. To reduce the search complexity, \textit{Darvish et al.}~\cite{darvish2021hierarchical} introduced a hierarchy of assemblies that are pre-computed offline and integrated with the first-order logic 
task description. Nevertheless, AND/OR graphs cannot efficiently model the subassembly's temporal constraints, such as parallelism. All assemblies are planned only in sequential order. Therefore, to achieve simultaneous executions, a dedicated task scheduler is required.
Other well-known graph-based task models are represented by Petri Nets~\cite{chen2014optimal,casalino2019optimal}, Hierarchical Task Networks 
~\cite{caccavale2017flexible}, Hierarchical Finite State Machines 
~\cite{nguyen2013rosco}, and Behavior Trees\cite{paxton2017costar}. 
In~\cite{chen2014optimal} a generalized stochastic Petri Net 
models all the possible actions required during the assembly task in single-human, multi-robot assembly cells. 
\textit{Casalino et al.}~\cite{casalino2019optimal}, instead, capitalizes on a partially controllable time Petri Net 
to model the unpredictability of human actions in human-robot collaboration. 
Behavior Trees (BTs) instead represent a planning method capable of modeling complex behaviors of autonomous agents. Due to their characteristics in terms of abstraction, modularity, and usability~\cite{biggar2022modularity}, they have been used in industrial settings to create and monitor task plans for industrial robots~\cite{paxton2017costar}. Despite the high potential of the formalism, in its current state, it does not provide the flexibility to adapt to human behavioral variability. 
Similarly to AND/OR graphs, 
Hierarchical Task Networks and Hierarchical Finite State Machines are planning methods based on action decomposition. 
In \cite{caccavale2017flexible} the Hierarchical Task Network models a structured cooperative task and reacts to unexpected events and behaviors. 

\begin{figure*}[t!]
    \centering
    \includegraphics[trim=0.0cm 0.0cm 0cm 0.cm,clip,width=\textwidth]{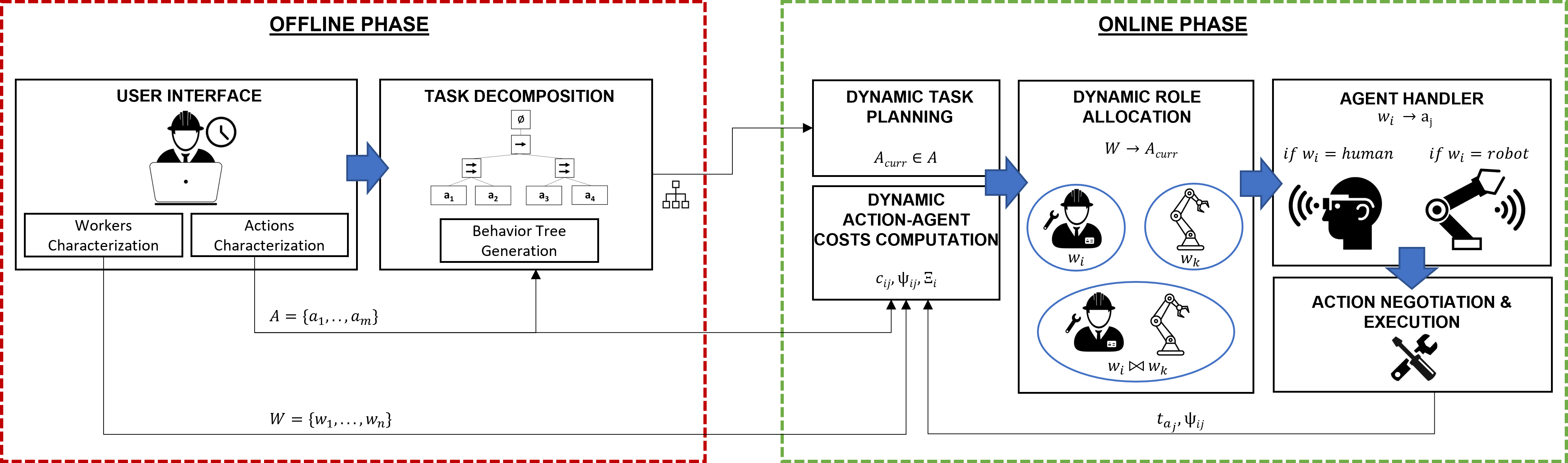}
    \caption{Architecture of the proposed framework. In order to model a task, action and worker features are user-defined offline, and the plan is built by decomposing a job into its atomic actions. In an online phase, the \textit{planner} follows the plan structure until one or more actions can be executed at the same time. To understand which worker is in charge of them, the \textit{Role Allocator} optimizes the worker-action suitability, represented by the dynamically computed costs, designing cooperations or collaborations between agents.} 
    \label{fig:architecture_scheme}
    \vspace{-3mm}
\end{figure*}

In addition, allocation methods based on decision-making processes have the advantage of being able to integrate multiple criteria and quickly obtain the desired allocation. For these reasons, such methods are particularly suitable in the context of dynamic role allocation with re-planning capability. Similar approaches are presented in \cite{el2022hierarchical, tsarouchi2017human}, where the task is represented through a 
Hierarchical Finite State Machine, while the allocation is done through a decision-making process that considers agent features such as availability, capability, workload, and performance. The re-scheduling capability of the latter is presented in~\cite{nikolakis2018dynamic}, where the multi-criteria decision-making framework computes offline task allocation and online reschedules when unexpected events occur. 
Ergonomic aspects are introduced by \textit{El Makrini et al.}~\cite{el2019task}. While these methods assume that the team is composed only of a single human and a single robot in synchronized processes, in~\cite{bruno2018dynamic} tasks can be assigned to both agents at the same time, allowing proper collaborative roles to be assigned. Yet, the generalization of such methods to a team of arbitrary size is nontrivial.
A schematic comparison of the literature review is available in \autoref{table:literature_comparison}\footnote{N/D in the table entries stands for no data.}.

%% file: sections/contribution.tex
The novelty of the proposed architecture in \autoref{fig:architecture_scheme} consists in combining, for the first time, BTs theory as a task scheduler with a MILP-based role allocation method that allows, within the same formulation, individual and collaborative roles.
BTs provide a simple yet effective way~\cite{paxton2018evaluating} to schedule task actions following the plan's constraints and broadcast their execution to the designated agent. This was possible by means of a set of novel custom BT nodes that could be easily deployed with BT concepts. 
Unlike conventional approaches, where each agent's behavior is governed by its BT-based planner, here the BT models the task rather than the agent.
The goal of the task planner is also to decompose the complex task into its atomic actions, ensuring that the temporal dependencies between actions (sequence and parallel) are observed. In this way, the complexity of the role allocation problem is reduced to different suboptimal but computationally less expensive problems without any constraints generated by the plan. In particular, optimality depends on the trade-off between the depicted costs (modeling agents' features) and the agents' availability. This means that, in some circumstances, the system might consider it worthwhile to wait for a currently unavailable agent, which could, alone or collaboratively, perform the selected action more effectively. The usage of a task-agnostic planner and the generality of the MATA through MILP provide the architecture with high scalability and modularity, in terms of the categories of tasks that can be modeled and the number of agents that can be handled.
The framework aims to generate plans that promote two different levels of engagement between the agents of the team, which are \textit{cooperation} and \textit{collaboration}. In particular, with cooperation, we refer to plan execution where agents work simultaneously in a shared workspace and have a common supreme goal, but do not work at the same time on the same workpiece, while with collaboration, which is the highest level of engagement, agents are physically coupled, balancing the efforts to achieve a shared goal by working on the same workpiece.
Collaborative roles evaluated in the work are $2$-agents combinations of the whole team (robot-robot, human-human, and human-robot). However, the formulation does not differ in the case of a $n$-agents combination, with $2 < n \le N$. 

The potential of the architecture is evaluated in three different steps. First, the computational complexity of the proposed methodology is investigated and compared with a similar State-of-the-Art (SoA) method based on AND/OR graphs~\cite{johannsmeier2017hierarchical,darvish2018flexible,darvish2021hierarchical}.
The rationale behind our choice to compare our architecture with AND/OR graphs lies in the disparity in methodologies. While AND/OR graphs rely purely on the graph, solving the planning problem as a graph search, our hybrid approach mixes the advantages of graphs in terms of task modeling while exploiting the computational efficiency of optimization methods for small-sized problems.
In this evaluation, we progressively increase the number of workers and actions. Second, a simulated experiment is exploited to compare the proposed variations of the proposed architecture in terms of the planned sequence of actions and the allocated actions. In this experiment, the job structure and the action-related costs, which are representative of the action duration (minimum time duration solution), were designed to highlight the main differences between the variations, demonstrating the parallelization capability of the algorithm with a nonspecific task example. Finally, the architecture was tested in a realistic scenario with an assembly task of a table-like structure performed collaboratively by a subject with a Franka Emika Panda. In this step, three different sets of experiments are performed. In a preliminary pilot test, we evaluated the efficiency of the proposed AR-based human-robot interface. Three different communication protocols are compared by the total execution time of the job. An additional experiment demonstrated how it is possible to include additional optimization principles by modifying the cost functions. In particular, the allocation results of the minimum job duration approach are compared with those of a psychologically safe human-robot interaction approach, where the robot-worker distance is combined with the action duration. The last experiment evaluated the acceptability of the overall architecture with a multi-subject subjective evaluation.

A preliminary version of this architecture was introduced in~\cite{fusaro2021integrated} where BTs were first presented as a method to plan subtasks that could later be allocated by means of different MILPs. However, this version of the method could not deal with i) a general structure of tasks (and hence allowing the maximum degree of action parallelization) and ii) proper human-robot collaborative action executions, where only single agent actions are envisioned (cooperation). Furthermore, in this work, a novel AR-based user interface to enable two-way human-system communication is introduced, a comprehensive characterization of the planning and allocation algorithm computational complexity is presented, and a comparison with SoA methods provides extensive validation of the method in a realistic scenario.

%% file: sections/methodology.tex
A schematic representation of the proposed architecture is shown in \autoref{fig:architecture_scheme}. The architecture requires an offline initialization phase, where the problem components are defined. In this work, human and robotic agents are generically called workers, while actions represent the atomic components of a complex task, which is referred to as job.
In particular, we assume that, given a job, it is always possible to break it down hierarchically into its atomic components, which are called actions. An action is defined as the intentional intervention of an agent directed towards objects, other agents, or the surrounding environment. 
For example, the assembly of a table (i.e., the job) can be decomposed first in the assembly of the legs with the tabletop. In their turn, each leg assembly can correspond to the sequence of leg positioning on the tabletop (action 1) and then the fastening of the two pieces (action 2).
Instead, in the manuscript, a task is used in a generic way to group single actions into a job.

Action-related and worker-related information is used to define the suitability of the worker for the action and is exploited to dynamically compute costs in the MILP formulation. Moreover, the job should be provided in BT form, 
and the actions that the robotic workers are capable of executing should be pre-programmed. In the online phase, the BT communicates to MATA the feasible actions that should be assigned according to the plan. For those actions, the worker-to-action suitability is evaluated and embedded in the MATA as problem costs. Once the MILP solver returns the worker-action pairs, the Agent Handler ensures that each action and allocation mode is broadcast to the designated worker. If the worker is a human, the handler communicates the action to the worker's AR interface, enabling a negotiation in which the worker can accept or refuse to execute the action; otherwise, if it is a robot, a combination of action primitives is activated.
It is important to highlight that a rejection of the human in the negotiation phase does not imply a direct allocation of the action to another worker (e.g., the robot). In our architecture, each allocation is the outcome of the optimization procedure, which evaluates all the other workers and finds the most suitable, which might be another human or one robot on the team. Thus, after a rejection, the role allocation method updates worker-action costs (which include the negation of the worker) and computes a new assignment for the available agents. If the same action is assigned to the same agent again, the action will be directly communicated to the worker without negotiation.

\subsection{Background on Behavior Trees}\label{ssec:background}
\input{sections/background_bt}

\subsection{Modeling Human-Robot Collaborative Plans}\label{ssec:hrc_model}
\input{sections/hrc_model}

\subsection{Role Allocation}\label{ssec:role_allocation}
\input{sections/role_allocation}

\subsection{User Interface}\label{ssec:ui}
\input{sections/user_interface}

%% file: sections/background_bt.tex
\begin{table}[!t]
\title{BT node types and return status.}
\begin{center}
\resizebox{\columnwidth}{!}{\begin{tabular}{|P{.2\columnwidth}|P{.2\columnwidth}|P{.2\columnwidth}|P{.2\columnwidth}|P{.2\columnwidth}|}
\hline
\multirow{2}{*}{\textbf{Type of Node}} & \multirow{2}{*}{\textbf{Symbol}} & \multirow{2}{*}{\textbf{Success}} & \multirow{2}{*}{\textbf{Failure}} & \multirow{2}{*}{\textbf{Running}} \\[3mm]
\hline
\end{tabular}}

\vspace*{0.1 cm}

\resizebox{\columnwidth}{!}{\begin{tabular}{|P{.2\columnwidth}|P{.2\columnwidth}|P{.2\columnwidth}|P{.2\columnwidth}|P{.2\columnwidth}|}
\hline
\multirow{3}{*}{Sequence} & \multirow{3}{*}{$\rightarrow$} &\multirow{2}{*}{All children} \multirow{2}{*}{succeed}  & \multirow{3}{*}{One child fails} & One child returns Running \\
\hline
\multirow{3}{*}{Fallback} & \multirow{3}{*}{?} & \multirow{2}{*}{One} \multirow{2}{*}{child succeeds} & \multirow{2}{*}{All children} \multirow{2}{*}{fail} & One child returns Running \\
\hline
\multirow{2}{*}{Decorator} & \multirow{2}{*}{$\diamondsuit$} & \multirow{2}{*}{Custom} & \multirow{2}{*}{Custom} & \multirow{2}{*}{Custom} \\[3mm]
\hline
\multirow{2}{*}{Parallel} & \multirow{2}{*}{$\rightrightarrows$} & $\geq M$ children succeed & $> N - M$ children fail & \multirow{2}{*}{else}\\
\hline
\end{tabular}}

\vspace*{0.1 cm}

\resizebox{\columnwidth}{!}{\begin{tabular}{|P{.2\columnwidth}|P{.2\columnwidth}|P{.2\columnwidth}|P{.2\columnwidth}|P{.2\columnwidth}|}
\hline
\multirow{2}{*}{Condition} & \multirow{2}{*}{\tikz \draw (0,0) ellipse (4pt and 2pt);} & \multirow{2}{*}{True} & \multirow{2}{*}{False} & \multirow{2}{*}{Never}\\[3mm]
\hline
\multirow{2}{*}{Action} & \multirow{2}{*}{\tikz \draw (0,0) rectangle (8pt,4pt);} & Upon completion  & Impossible to complete & During execution\\
\hline
\end{tabular}}
\end{center}
\caption{BT node types and return status.}
\label{table:BT_nodes}
\vspace{-4mm}
\end{table}

In the proposed architecture, Behavior Trees (BTs) are exploited as a task model. In particular, they allow the decomposition of the job into a group of atomic actions and the dynamic scheduling of the atomic actions according to the constraints of the job.


A Behavior Tree (BT) is defined as a rooted tree, where each adjacent pair of nodes, referred to as \textit{parent} and \textit{child} node, is connected by directed edges.
It is characterized by internal nodes responsible for regulating control flow and leaf nodes designated for the execution of actions or the evaluation of conditions.
The root node initiates the sequential execution of its child nodes by broadcasting a signal termed as \textit{tick} throughout the tree structure, from its bottom left to its bottom right. Upon receiving the tick, the child nodes promptly relay a status update to their parent node: if the child is actively executing, it furnishes a 'RUNNING' status; conversely, upon successful completion of the node's execution, it imparts a 'SUCCESS' status, while in the event of failure, a 'FAILURE' status is communicated. Within the BT formulation, two fundamental node types emerge: control flow nodes and execution nodes. The former type, which has at least one child, includes four established categories, namely Sequence, Fallback, Parallel, and Decorator, while the latter, which corresponds to the leaves of the tree, contains two categories, namely Action and Condition. The node types, distinguished by their respective symbols and the ensuing status in each scenario, are encapsulated in \autoref{table:BT_nodes}. The canonical formulation of a BT has been artfully engineered to oversee the behavioral dynamics of agents, facilitating the real-time reactive formulation of executable tasks~\cite{colledanchise2018behavior}. This characteristic favors the generation of diverse behaviors tailored to meet different conditions on the fly.

Thanks to the abstraction of the planning method, the BT can be used for a general class of industrial processes and easily adapted to embed mixed human-robot teams and collaborations between agents. The first attempt at extending the paradigm of a single robot BT to a multi-robot BT was proposed in ~\cite{colledanchise2016advantages}, where a systematic procedure to map a single robot BT to a set of BTs, capable of achieving the same goal as the starting one, is proposed. The manuscript well explains the concept of parallelization of tasks, which is also exploited in our methodology, and introduces the presence of a simplified Task Assignment node. However, the latter does not envison the presence of humans in the team or the collaboration between teammates.

%% file: sections/hrc_model.tex
\begin{figure}[t]
    \centering
    \begin{subfigure}[t]{\columnwidth}
        \centering
        \includegraphics[trim=0cm 0.0cm 0cm 0cm,clip,width=\linewidth]{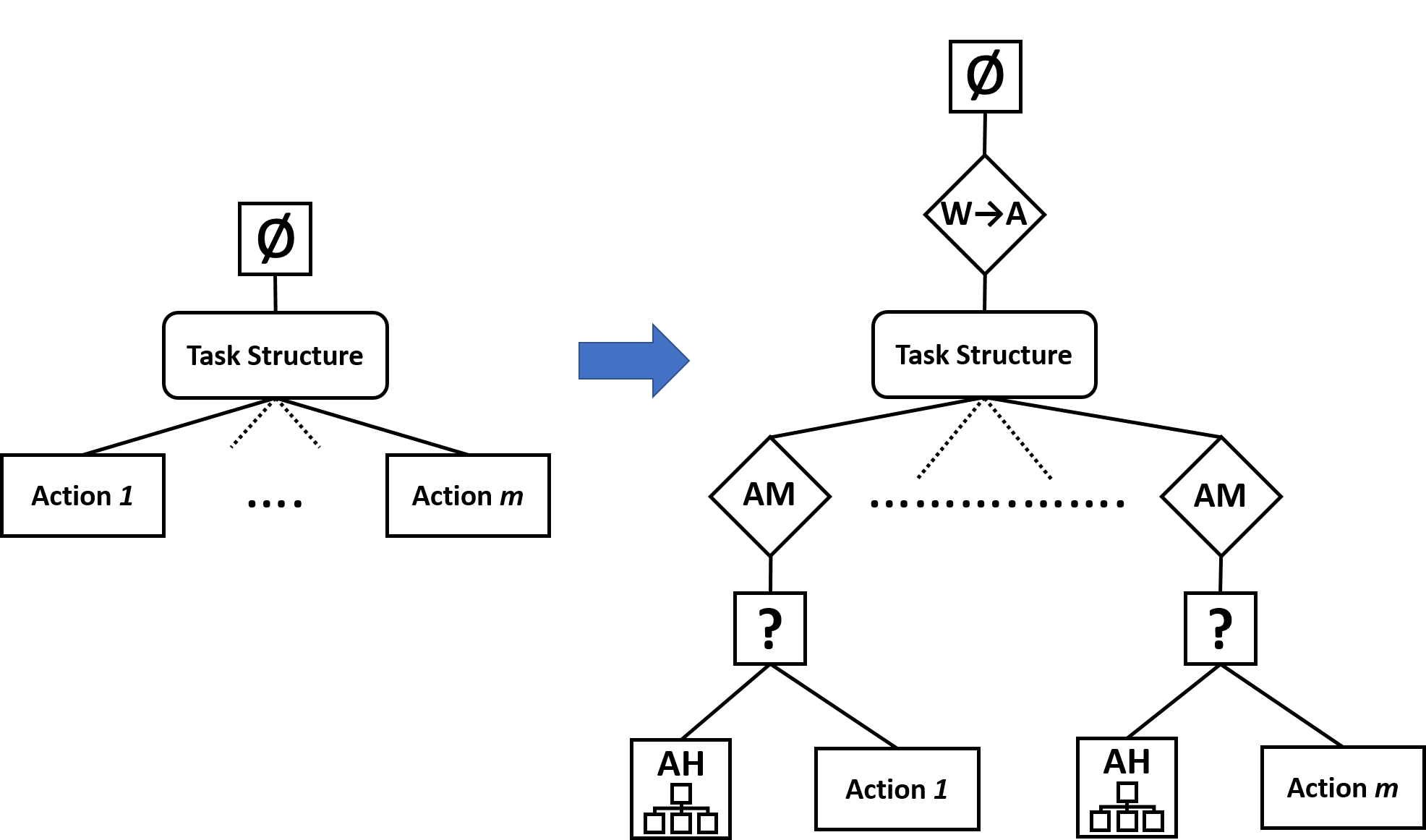}
        \caption{Comparison between the standard BT structure to model a job made of m actions (left) and the new BT design (right) that includes the role allocation process for the same job.   }
        \label{fig:BT_structure}
    \end{subfigure}
    \begin{subfigure}[t]{\columnwidth}
        \centering
        \includegraphics[trim=0mm 0.0cm 0mm 0cm,clip,width=0.6\linewidth]{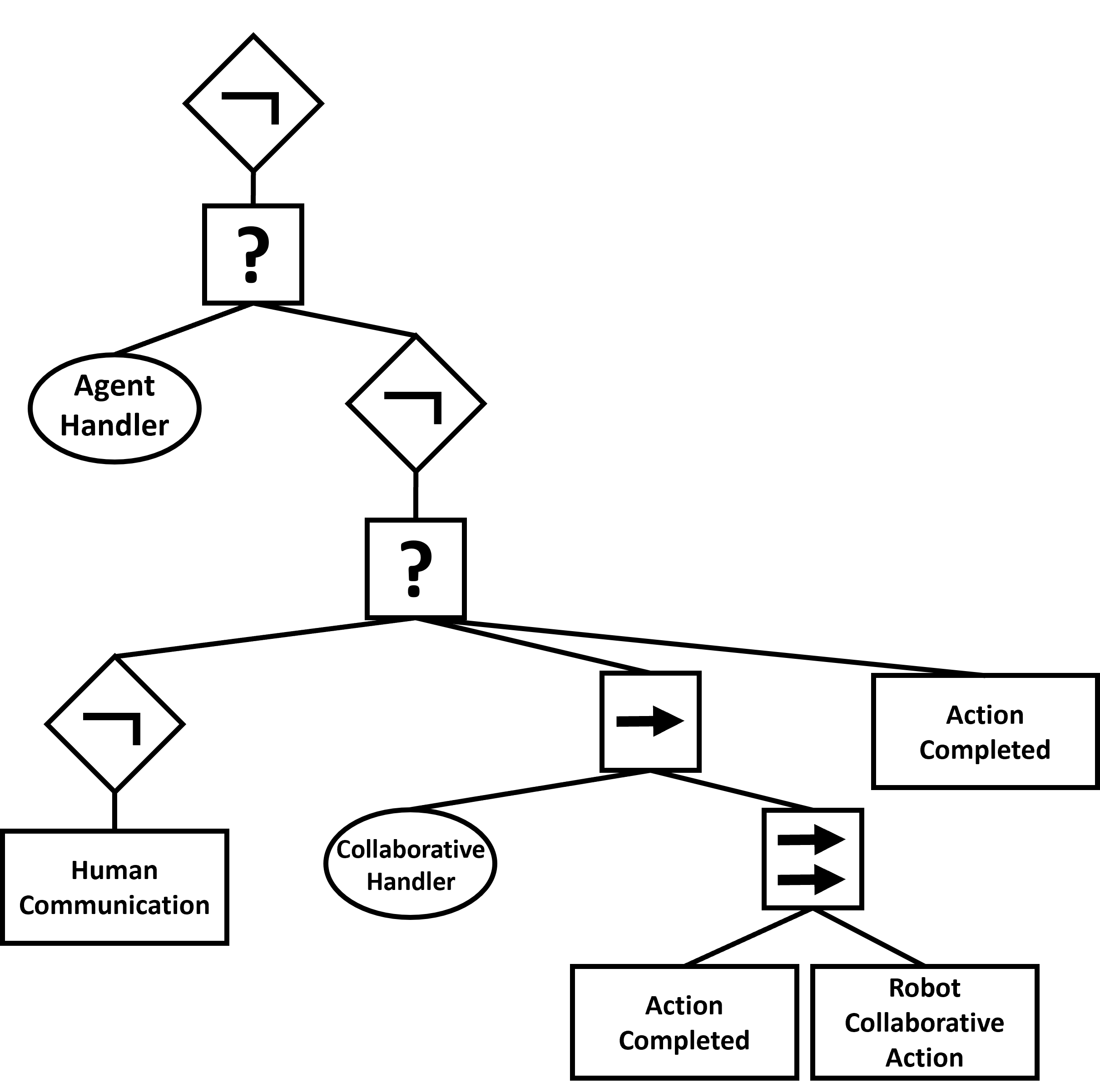}
        \caption{Explosion of the Allocator Helper subtree.}
        \label{fig:AH_subBT_structure}
    \end{subfigure}
    \caption{Formulation of the role allocation problem using a custom structure of novel BT nodes. In the new design, the modularity of the BT paradigm is maintained, since it requires one single node for the allocation and a modular structure to communicate the action to the allocated agent.}
    \label{fig:robots}
    \vspace{-2mm}
\end{figure}

To further integrate the MATA problem into the scheduling process, the standard BT structure and functionalities are extended. The idea was to keep the same theory and rules of the standard BT to generate robotic plans and to add novel components, i.e., BT nodes, dedicated to the role allocation and scheduling of human-robot systems. In this way, the task model could be built using the same skills and expertise required by standard BTs~\cite{paxton2018evaluating}.  
The new plan structure adds three main components: a \textit{Role Allocator} node $W \! \rightarrow \! A$, and, for each action, an \textit{Allocation Manager} node and \textit{Allocator Helper} subtree. In \autoref{fig:BT_structure} the new task planner is compared to the standard BT-based planner. The Allocator Helper subtree, in turn, makes use of four different nodes, i.e. the \textit{Agent Handler}, the \textit{Human Communication}, the \textit{Collaborative Handler} and the \textit{Action Completed} (see \autoref{fig:AH_subBT_structure}).
The node functions are the following: the Agent Handler selects the right branch of the tree according to allocation results; the Human Communication broadcasts the allocation result to the right human worker; the Collaborative Handler checks and activates collaborative action executions according to allocation results; and the Action Completed is used to acknowledge the achievement of the action by the human.
Finally, the new structure calls for dedicated collaborative action nodes besides the standard robot action node, which allows, if expected by the Role Allocator, collaborative action executions.



\subsubsection{Role Allocator Node (Algorithm~\ref{algo:RoleAllocator})}
This node is the child of the root node. It is defined as a decorator node and, hence, it has only one child.
The child of this node is the Task Structure\footnote{Note that the Task Structure is not a BT node, but it represents a part of the tree related to the deployment of the task with standard BT nodes.}, which is the hierarchical structure of the plan and contains action preconditions and plan constraints. 
The goal of the node is to compute the agent-action related costs, solve the role allocation problem 
and generate an output based on the result, which includes an allocated action for each agent.
The evaluated actions are only a subset of all the job actions, i.e., the ones feasible according to the plan constraints. To do so, the node reads the information about the ticked actions from the input data set by the Allocator Manager node, as well as the agents' information for the cost computation. 
For this reason, the first time the node is inspected, it only ticks the child since there are no actions to be allocated (no Allocator Manager ticked). From this moment on, the node is RUNNING and loops on solving the MATA at a desired frequency ($100\;Hz$ in our case), until all the actions are completed.



\alglanguage{pseudocode}
\begin{algorithm}[!ht]
\small
\caption{Tick() function of the Role Allocator node.}
\label{algo:RoleAllocator}
\begin{algorithmic}[1]
\Procedure{$\textproc{RoleAllocator::tick}$}{$ $}
\State $getInput(acts\_to\_be\_allocated)$
\If {$acts\_to\_be\_allocated \neq 0$}
    \State $[W,\;A] = Allocate()$
    \State $setOutput([W,\;A])$
\EndIf
\State $child\_status \gets child.Tick()$
\If {$child\_status ==$ \emph{FAILURE}} 
    \State \textbf{return} \emph{FAILURE}
\ElsIf {$child\_status ==$ \emph{SUCCESS}} 
    \State \textbf{return} \emph{SUCCESS}
\EndIf
\State \textbf{return} \emph{RUNNING}
\EndProcedure
\Statex
\end{algorithmic}
\vspace{-2mm}
\end{algorithm}

\subsubsection{Allocator Manager AM Node (Algorithm~\ref{algo:AllocatorManager})}
While in the standard BT the robotic actions are linked directly to the Task Structure, here they are children of an Allocator Manager 
node (one per action).
As the job advances during action execution, the BT propagates the tick signal through the Allocator Manager. It is defined as a decorator node, and it is in charge of managing the information required by the Role Allocator related to the linked action and the possible rejection of the allocated human for the same action. 
Once it has received the allocation and checked that its related action is allocated, it verifies that the assigned agent is available. In this case, the tick signal is propagated to the child.
In case the child subtree returns \textit{SUCCESS}, the node controls if the action was rejected by the human. If so, the node waits for the results of a new allocation; otherwise, it returns \textit{SUCCESS}.

\alglanguage{pseudocode}
\begin{algorithm}[!ht]
\small
\caption{Tick() function of the Allocator Manager node.}
\label{algo:AllocatorManager}
\begin{algorithmic}[1]
\Procedure{$\textproc{AllocatorManager::tick}$}{$ $}
\State \textproc{add}($child\_action$ \textproc{in} $acts\_to\_be\_allocated$)
\State $setOutput(acts\_to\_be\_allocated)$
\State $getInput([W,\;A])$
\For{$[w,\;a]$ \textbf{in} $[W,\;A]$}
    \If {$a == child\_action$}
        \If{$w.available$} 
            \State $action\_allocated =$ \textbf{true}
        \Else
            \State $action\_allocated =$ \textbf{false}
        \EndIf
        \State \textbf{break}
    \EndIf
\EndFor
\If {$action\_allocated$}
    \State $child\_status \gets child.Tick()$
    \If {$child\_status ==$ \emph{SUCCESS}} 
        \State $getInput(actions\_rejected)$
        \If {$child\_action$ \textproc{in} $actions\_rejected$}
            \State \textproc{erase}($child\_action$ \textproc{in} $actions\_rejected$) 
            \State $action\_allocated =$ \textbf{false}
        \Else
            \State \textbf{return} \emph{SUCCESS}
        \EndIf    
    \ElsIf {$child\_status ==$ \emph{FAILURE}} 
        \State \textbf{return} \emph{FAILURE}
    \EndIf
\EndIf
\State \textbf{return} \emph{RUNNING}
\EndProcedure
\Statex
\end{algorithmic}
\vspace{-2mm}
\end{algorithm}

\subsubsection{Agent Handler AH Node (Algorithm~\ref{algo:AgentHandler})}
The \textit{Agent Handler} node is responsible for directing the action to the correct agent according to the MATA results by activating one of the two different branches of the 
Allocator Helper subtree. Specifically, it is defined as a condition node, and 
if the assigned agents for the action include one or more humans (both alone or in collaboration with other types of agents), the node returns \textit{FAILURE} and so it propagates the signal through the subtree. On the other hand, if the allocated agents are only robotic, the Agent Handler returns \textit{SUCCESS} and hence the signal is propagated through the Robot Action nodes.

\subsubsection{Human Communication Node (Algorithm~\ref{algo:Communication})}
Once the Agent Handler selects the branch based on the allocation result of that specific action, if the allocated agents include at least a human, the Human Communication node is ticked. It is defined as an action node, and it broadcasts the allocated action to the human agent. As a matter of practice, it transfers the information to the \textit{User interface} (\autoref{ssec:ui}).
The node sends a request to the physical device that is in charge of notifying the human of the assignment, waits until the response of the negotiation is sent by the human, and sets the status of the agent as not available to avoid multiple actions being allocated to the same agent. It returns the corresponding value: \textit{SUCCESS} if the action is accepted and hence erased from the list of the actions that have to be allocated, while \textit{FAILURE} if rejected. Moreover, if the request for that action is declined by the human, the node updates the list of the rejected actions by the specific agent, which is required by the 
Allocator Manager updates the related agent-action costs (the procedure will be explained later on in \autoref{par:preference_cost}) and sets the status of the agent as available.


\alglanguage{pseudocode}
\begin{algorithm}[!ht]
\small
\caption{Tick() function of the Communication node.}
\label{algo:Communication}
\begin{algorithmic}[1]
\Procedure{$\textproc{Communication::tick}$}{$ $}
\State $getInput([W,\;A])$
\For{$[w,\;a]$ \textbf{in} $[W,\;A]$}
    \If {$w.type ==$ HUMAN $\lor$ $w.type ==$ COLLABORATIVE}
        \State \textproc{send}($request$ \textproc{for} $a$)
        \State $w.available$ = \textbf{false}
        \If {$response ==$ accepted}
            \State \textproc{erase}($child\_action$ \textproc{in} $acts\_to\_be\_allocated$)
            \State $setOutput(acts\_to\_be\_allocated)$
            \State \textbf{return} \emph{SUCCESS}
        \ElsIf {$response ==$ rejected}
            \State \textproc{add}($child\_action$ \textproc{in} $actions\_rejected$)
            \State $setOutput(actions\_rejected)$
            \State $updateCost([w,\;a])$
            \State $w.available$ = \textbf{true}
            \State \textbf{return} \emph{FAILURE}
        \EndIf
    \EndIf
\EndFor
\EndProcedure
\Statex
\end{algorithmic}
\vspace{-2mm}
\end{algorithm}

\subsubsection{Collaborative Handler Node (Algorithm~\ref{algo:CollaborativeHandler}) }
The Collaborative Handler node checks if the action is allocated to mixed teams, i.e., the action has to be accomplished in collaboration between human and robotic agents. Hence, it is defined as a condition node which, 
returns \textit{SUCCESS} if the action requires collaboration, \textit{FAILURE} otherwise.

\subsubsection{Action Completed Node (Algorithm~\ref{algo:ActionCompleted})}
Action Completed is defined as an action node that corresponds to a human execution of the action. In practice, once ticked, the node sends a request to the user device, which has to respond when the action is ended. Hence, the node keeps running until positive feedback is sent from the human. Once received, it sets the agent as available and returns \textit{SUCCESS}. 

\subsubsection{Robot Action Nodes}
The Robot Action nodes represent the physical implementation of the actions that the robot can perform. For simplicity, we assume that robotic actions are always successful, and the SUCCESS response is sent when they are completed. However, fault recovery strategies can be included here thanks to the BT formalism~\cite{colledanchise2016advantages}.
They can be assigned by the role allocator node. 
Examples of potentially assignable actions are \textit{insert object} or \textit{move object}.
In this manuscript, Robot Action nodes are predefined compositions (sequences or parallels) of primitives. While actions are job-specific, primitives are general and allow for the generation of particular job actions.
Examples of primitives are point-to-point motion, force exertion, the activation of a gripper, etc. Each primitive requires some specification data (for example, the target pose of the motion).
An example of the structure of the action \textit{move object} is given as the sequence of:
\begin{enumerate}
    \item Move to the object pose (MOVE);
    \item Close the gripper (GRASP);
    \item Move to the target pose (MOVE);
    \item Open the gripper (GRASP).
\end{enumerate}
The solution above explains how to generate hierarchical autonomous actions from primitives. However, the architecture also envisions collaborative human-robot behaviors, which often differ from autonomous ones.
However, the strategy to generate collaborative actions by combining different robot primitives is analogous.
Moreover, controller switching is also considered a primitive action.
A practical example, based on our strategy, is given in the \autoref{ssec:real_exp}.

%% file: sections/role_allocation.tex
\subsubsection{Problem Statement}
\label{sssec:problem}
Before formalizing the MATA problem, we need to specify the main components of the problem.
Following the formalism introduced in~\cite{lamon2019capability}, we consider a mixed human-robot team of workers, or agents, $W=\{w_1,...,w_n\}$, with $|W| = N$. The goal is to complete a general single job $Q$ that could be further decomposed into sequential and parallel actions, $A=\{a_1, .. ,a_m\}$, $|A| = M$. The set of actions that can be executed by each worker $w_i \in W$ are $A_i=\{a_{i1}, .. ,a_{io}\}$, $|A_i| = O$, where $A_i \subseteq A$. The goal is to obtain the allocation of an worker $w_i$ to each of the actions he/she/it is able to execute $a_{ij}$, which here is denoted by $w_i \rightarrow a_{ij}$ (read worker $i$ is assigned to action $j$). The set of worker-action allocations is denoted $W \rightarrow A$.

\subsubsection{Combinatorial Optimization Model for Cooperative Role Allocation}
\label{sssec:coop_model}
In this work, the MATA problem is formalized as a Mixed-Integer Linear Program. The most general scenario of MATA problems in human-robot collaboration, according to 
the taxonomy in ~\cite{korsah2013comprehensive} is characterized by multi-task agents, i.e., agents that can execute multiple tasks simultaneously, multi-agent tasks (tasks that require multiple agents), and finally, time-extended assignments, i.e., the allocation also considers future allocations. Such a problem is NP-hard~\cite{gombolay2018fast} and cannot be easily formulated through common combinatorial models present in \cite{korsah2013comprehensive}. 
However, in our framework, thanks to the decomposition of the task that BTs are able to achieve, the problem is extremely simplified. First, each worker, by definition, can perform only one action at a time; second, in a cooperative scenario, each action requires a single worker\footnote{Collaborative tasks, that require more than a single worker at the same time, will be addressed in~\autoref{sssec:collab_model}.}. 
By doing so, we have removed all the \textit{complex} and \textit{cross-schedule} dependencies, i.e., the effective cost of a worker for an action and the allocation constraints do not depend on the schedules of other workers.
In practice, the Role Allocator node deals each time only with the allocation of a sub-set of actions that, according to the progress of the job schedule represented by the BT, have as constraints, in the worst-case scenario, \textit{intra-schedule} dependencies. This means that the worker's cost for an action depends on the other actions the worker is performing.
Hence, the problem of allocating $L$ actions, where $L \le M$, to $N$ workers, can be formalized in the following way:

\vspace{1mm}
Minimize
\begin{equation}\label{eq:role_allocation_cost}
   \sum_{w_i\in W}\sum_{a_j\in A}( c_{ij} + \xi_i)x_{ij} 
\end{equation}

Subject to
\begin{subequations}
\begin{align}
    x_{ij} & \in\{0,1\} & \forall\; w_i \in W, \forall\; a_j \in A \label{eq:role_allocation_constr_bin} \\
    x_{ij} & = 0 & \text{ if } a_j \notin A_i 
    \label{eq:role_allocation_constr_not_capable} \\
    \sum_{w_i \in W} x_{i j} & \le 1 & \forall\; a_j \in A
    \label{eq:role_allocation_constr_one_task} \\
    \sum_{a_j \in A} x_{i j} & \le 1 & \forall\; w_i \in W
    \label{eq:role_allocation_constr_one_agent}\\
    \sum_{w_i \in W} \sum_{a_j \in A} x_{i j} & = \min{(L, N)}
    \label{eq:role_allocation_constr_number_alloc}\\
    \sum_{w_i \in W} \sum_{a_j \in A} t_{i j} x_{i j} & \leq T_{k} & \forall\; k \in K 
    \label{eq:role_allocation_constr_budget}
\end{align}
\label{eq:role_allocation_constr}
\end{subequations}

\noindent where $c_{ij}$ represents the cost related to an worker $w_i$ in executing the action $a_j$, $\xi_i$ is the availability cost function (\autoref{par:availability_cost}), and the $x_{ij}$ represents the $L \times N$ binary optimization variables of the problem, where $x_{ij} = 1$ means that the worker $i$ is assigned to action $j$ ($w_i \rightarrow a_{j}$).

The problem constraints are exploited to ensure the feasibility of the solution:
\begin{itemize}
    \item[\eqref{eq:role_allocation_constr_bin}] is representative of the binary nature of the variables;
    \item[\eqref{eq:role_allocation_constr_not_capable}] ensures that the solver does not allocates a worker to an action that is not capable of execution; 
    \item[\eqref{eq:role_allocation_constr_one_task}]\!\!\& \eqref{eq:role_allocation_constr_one_agent} ensure that to each worker only one action is allocated, and the same action is not allocated to multiple workers.
    \item[\eqref{eq:role_allocation_constr_number_alloc}] ensures that there are exactly a number of allocations equal to the number of workers $N$, in case $N < L$, where $L$ are the actions to be allocated and $L$ otherwise;
    \item[\eqref{eq:role_allocation_constr_budget}] is the budget constraint that allows further limits to the number of actions assigned to each worker, where $t_{ij}$ is the budget cost that $w_i$ would spend for $a_j$, and $T_k$ is the budget limit for the $K$ joint worker-action constraints. 
\end{itemize}

\subsubsection{Combinatorial Optimization Model for Collaborative Role Allocation}
\label{sssec:collab_model}
The main limitation of the allocation method in the previous subsection,
as well as most of SoA methods \cite{johannsmeier2017hierarchical, darvish2021hierarchical, pupa2021human, cramer2021probabilistic}, consists in the fact that it is suitable only for synchronized/cooperative tasks, where agents work simultaneously in the shared workspace, but it does not model properly human-robot collaborative tasks, i.e. tasks where agents work on the same workpiece at the same time, balancing their efforts to achieve a shared goal.

In practice, in our allocation method, we need to ensure that multiple workers can be assigned to the same action. To do so, we modified the formulation of the MILP in \eqref{eq:role_allocation_cost},\eqref{eq:role_allocation_constr} in the following way.
First, we replace the set of $N$ workers $W=\{w_1,...,w_n\}$ with an augmented set $W$ where all the possible combinations of $N$ workers are considered:
\begin{equation} \label{eq:W_workers}
    \begin{aligned} 
        W &= W_1 \cup W_2 \cup W_3 \cup ... \cup W_N= \\
        &= \{w_1,...,w_n\} \cup \{w_{1\Join2},...,w_{n-1\Join n}\} \cup \\ 
        & \cup \{w_{1\Join2\Join3},...,w_{n-2\Join n-1\Join n}\} \cup ... \cup \{w_{1\Join...\Join n} \} = \\
        &= \{w_1,...,w_n,w_{1\Join2},...,w_{n-1\Join n}, \\ & \qquad w_{1\Join2\Join3},...,w_{n-2\Join n-1\Join n},...,w_{1\Join...\Join n} \} 
    \end{aligned}
\end{equation}
where $w_{i\Join j}$ represent the collaboration between $w_i$ and $w_j$, 
$|W|= \sum_{i=1}^{N}{\binom{N}{i}} = P$, and $W_k$ is the subset of $W$ that contains the $k$-combination of the $N$ workers, $k$$=$$1,...,K,...,N$. 
This choice represents the fact that, in theory, all workers could collaborate to achieve each single action. In practice, in HRC, usually a small number of workers are paired. For this reason, without loss of generality, only the 2-combinations ($K=2$) of $W$ are considered:
\begin{equation} \label{eq:W_agents_2}
    \begin{aligned}
        W = W_1 \cup
        W_{2}=\{&w_1,...,w_n,w_{1\Join2},...,w_{n-1\Join n} \}
    \end{aligned}
\end{equation}
where now $|W|= \sum_{i=1}^{2}{\binom{N}{i}}=\sum_{i=1}^{N}{i} = (N+1) \dfrac{N}{2} = P$. For example, in case of a human-robot team composed of 2 workers, $w_{h_1}$ and $w_{h_2}$ and a robotic co-worker $w_{r_1}$, for a single action $a_1$, a total of $P=6$ allocation candidates are considered: $W=\{w_{h_1},w_{h_2},w_{r_1},w_{h_1\Join h_2},w_{h_1\Join r_1},w_{h_2 \Join r_1}\}$.
Thanks to this choice, the new allocation problem can be formulated as follows:

\vspace{1mm}
Minimize
\begin{equation}\label{eq:role_allocation_collab_cost}
    \sum_{w_i\in W}\sum_{a_j\in A}( c_{ij} + \psi_{ij} + \Xi_i)x_{ij} 
\end{equation}

Subject to
\begin{subequations}
    \begin{align} \label{eq:role_allocation_collab_constr}
    x_{ij} & \in\{0,1\} & \forall\; w_i \in W, \forall\; a_j \in A \\
    x_{ij} & = 0 & \text{ if } a_j \notin A \\
    \sum_{w_i \in W} x_{i j} & \le 1 & \forall\; a_j \in A \\
    \sum_{a_j \in A} x_{i j} & \le 1 & \forall\; w_i \in W \\
    \sum_{k=1}^{N} \sum_{w_i \in W_k} \sum_{a_j \in A} \theta_k x_{i j} &= \min{(L, N)} & \label{eq:role_allocation_collab_constr_number}\\
    \sum_{w_i \in W} \sum_{a_j \in A} \boldsymbol{\eta}_i^T \boldsymbol{e}_k x_{i,j} & \leq 1 & \forall\;k=1,...,N \label{eq:role_allocation_collab_constr_unicity}\\
    \sum_{w_i \in W} \sum_{a_j \in A} t_{i j} x_{i j} & \leq T_{k} & \forall\; k \in K 
    \end{align}
\end{subequations}
\noindent where now $c_{ij}$ represents the cost related to an worker $w_i$ or the collaboration of more workers (if $i=a\Join b$ and $w_a,w_b$ are two different workers) in executing the action $a_j$. Moreover, $x_{ij} = 1$ means that either a single worker $i$, or more workers in collaboration (if $i=a\Join  b$) are assigned to action $j$ ($w_i \rightarrow a_{j}$). 
Moreover, $\psi_{ij}$ is the preference cost (\autoref{par:preference_cost}), $\Xi_i$ is the availability cost function that accounts also for the availability of the collaboration:
\begin{equation} \label{eq:availability_collab}
    \Xi_i = \begin{cases}
    \xi_a, \quad &\text{if } i=a \qquad \quad  \textit{(single worker)}, \\
    \max{(\xi_a,\xi_b)}, &\text{if } i=a\Join b \quad \; \textit{(collaboration)}.
    \end{cases}
\end{equation}
where $\xi_a$ can be computed as explained in \autoref{par:availability_cost}.
The constraints are similar to those presented in \eqref{eq:role_allocation_constr}, but are adapted to consider the presence of collaborative solutions. Also in this case, if an action $a_j$ is not feasible collaboratively by workers $w_a$ and $w_b$  ($i=a\Join  b$), one just need to set $x_{ij}=0$ as constraint. Then:
\begin{itemize}
        \item[\eqref{eq:role_allocation_collab_constr_number}] ensures that there are exactly a number of allocations equal to the number of total workers $N$, in case $N < L$, where $L$ are the actions to be allocated and $L$ otherwise. In this case, since multiple workers can be allocated to the same action, we need to scale the contribution to the overall solution counter, from the value $k$ defined in the range $[1,N]$, to the value $\theta_k$ defined in the range $[1,\min(L,N)]$, computed as:
        \begin{equation} \label{eq:sol_count_collab}
            \theta_k =  \bigg\lfloor \dfrac{k-1}{N-1+\varepsilon}(\min{(L, N)}-1)+1 \bigg\rfloor.
        \end{equation}
        The $\varepsilon>0$, small enough, is introduced to avoid numerical issues when $N=1$.
        \item[\eqref{eq:role_allocation_collab_constr_unicity}] are $N$ constraints that secure that, if a worker is involved in a collaborative allocation, it cannot be selected for any other allocation (both as single worker and also in other collaborations), and vice-versa. To do so, for each worker, we associate the vector ID $\boldsymbol{\eta}_i$ with the $i$-th worker, where $\boldsymbol{\eta}_i$ $(\forall \;i = 1,\;...\:,P)$ is defined as follows: 
        \begin{align}
            \boldsymbol{\eta}_i = 
            \begin{cases}
                \boldsymbol{e}_k  &\text{if } 
                i = k
                \qquad \quad  \textit{(single worker)},\\
                {\boldsymbol{e}_j} + {\boldsymbol{e}_k} &\text{if }
                i = j \Join k
                \quad \; \textit{(collaboration)},
            \end{cases}
        \end{align}
        where $\boldsymbol{e}_k$ is a canonical vector of $\in\mathbb{R}^N$. In practice $\boldsymbol{\eta}_i$ is a vector with binary entries, with zeros except in the correspondence of the workers involved in the collaboration, where there are ones. This rule can be easily extended to a larger number of workers in collaboration. 
\end{itemize}



\subsubsection{Costs Design}
\label{sssec:costs_design}
In this work, the optimization costs are split into two main components: the worker-actions costs $c_{ij}$ and the worker-related availability cost $\xi_i$. The former should describe how capable a worker is in performing each action, considering the kino-dynamics features of the workers, the actions duration, the human ergonomics, etc. The smaller the cost, the higher the suitability. 
The latter, instead, is fundamental to optimize not only the worker-action fitness, but also in terms of action execution time. If used together, the solution represent the trade-off between minimizing the workers waiting times and maximizing their suitability to the action. While the abstraction of the framework allows us to benefit of different cost functions and the design of new suitability indicators is out of the scope of the paper, still we would like to provide some guidelines on the costs design. 

\paragraph{Designing the Worker-Action Costs}
\label{par:suitability_cost}
In this paragraph, we highlight different aspects that can be considered in cost design. However, the specific choice is task-dependent and beyond the goal of the paper. We refer to~\cite{lamon2019capability} for an in-depth analysis of the factors and design of costs in a role allocation problem.
The concept of capability derives from the utility value described by \textit{Gerkey and Matari\'c}, as the cost of executing an action~\cite{gerkey2003role}, estimated as a combination of different sensor information. In their setup, the team is composed of workers of the same type, and hence the utility depends mainly on the action. Typical production line indicators are process time~\cite{chen2014optimal}, additional investment, and process quality, estimated from work measurements techniques like MTM (\textit{method time measurement}) and RTM (\textit{robot time and motion}).
In mixed teams, such as in HRC, all these aspects, together with workers' characteristics, play a fundamental role in the cost that a worker pays to achieve an action. In a previous study, the authors investigated the measurable capabilities that both humans and robots share~\cite{lamon2019capability}. 
These dynamic capabilities can complement the static binary capabilities (capable/not capable) included as optimization constraints. 
Three indexes are presented, \textit{task complexity}, \textit{agent dexterity}, and \textit{agent effort}. While the first one models the feasibility of an action by the worker, the other two evaluate the kinematic and dynamic properties of workers. By including these workspace-dependent quantities in the problem, the strategies computed iteratively by the Role Allocator node will differ according to the worker state, time-variable plans, and collaboration strategies.
Nevertheless, such indexes are quite complex and require profound robotic knowledge (robotic and human modeling, inverse kinematics, and inverse dynamics) to be computed, and hence, might not be suitable for industrial settings. Furthermore, other indexes have already been applied in manufacturing, such as human ergonomics risk indicators, both posture-related (e.g., \textit{REBA}~\cite{hignett2000rapid}) and task-specific (e.g., \textit{WISHA}~\cite{waters1993revised} or \textit{EAWS}~\cite{schaub2013european}). 
Moreover, the cost choice could also affect the teamed performance by changing the robot's online behavior. Some works demonstrated that different factors, such as robot trajectories~\cite{dragan2013legibility,bekkis2020robot}, velocity or acceleration~\cite{kulic2007physiological,lagomarsino2022robot}, and human-robot distance~\cite{arai2010assessment,bergman2020close}, influence the collaboration in terms of stress and mental fatigue. 
For these reasons, costs could change dynamically to prevent those undesirable conditions. 
 
\begin{figure*}[!t]
     \centering
     \begin{subfigure}[b]{0.19\textwidth}
         \centering
    	 \includegraphics[trim=0.0cm 0.0cm 0.0cm 0.0cm,clip,width=3.2cm,height=2cm]{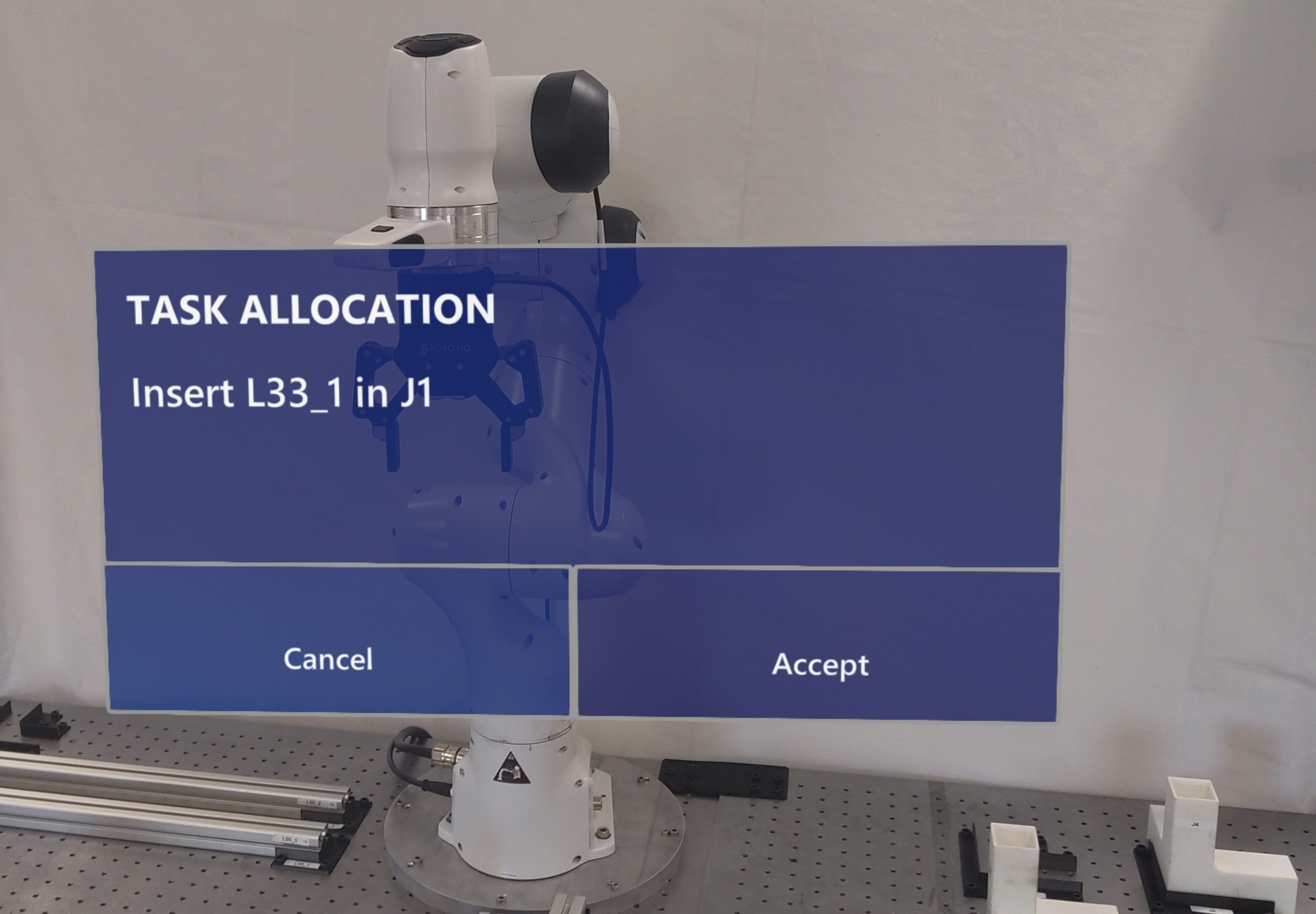}
	     \caption{Human allocation.}
	     \label{fig:UI_alloc}
     \end{subfigure}
     \hfill
     \begin{subfigure}[b]{0.19\textwidth}
         \centering
	     \includegraphics[trim=10cm 8.7cm 0.0cm 15.0cm,clip,width=3.2cm,height=2cm]{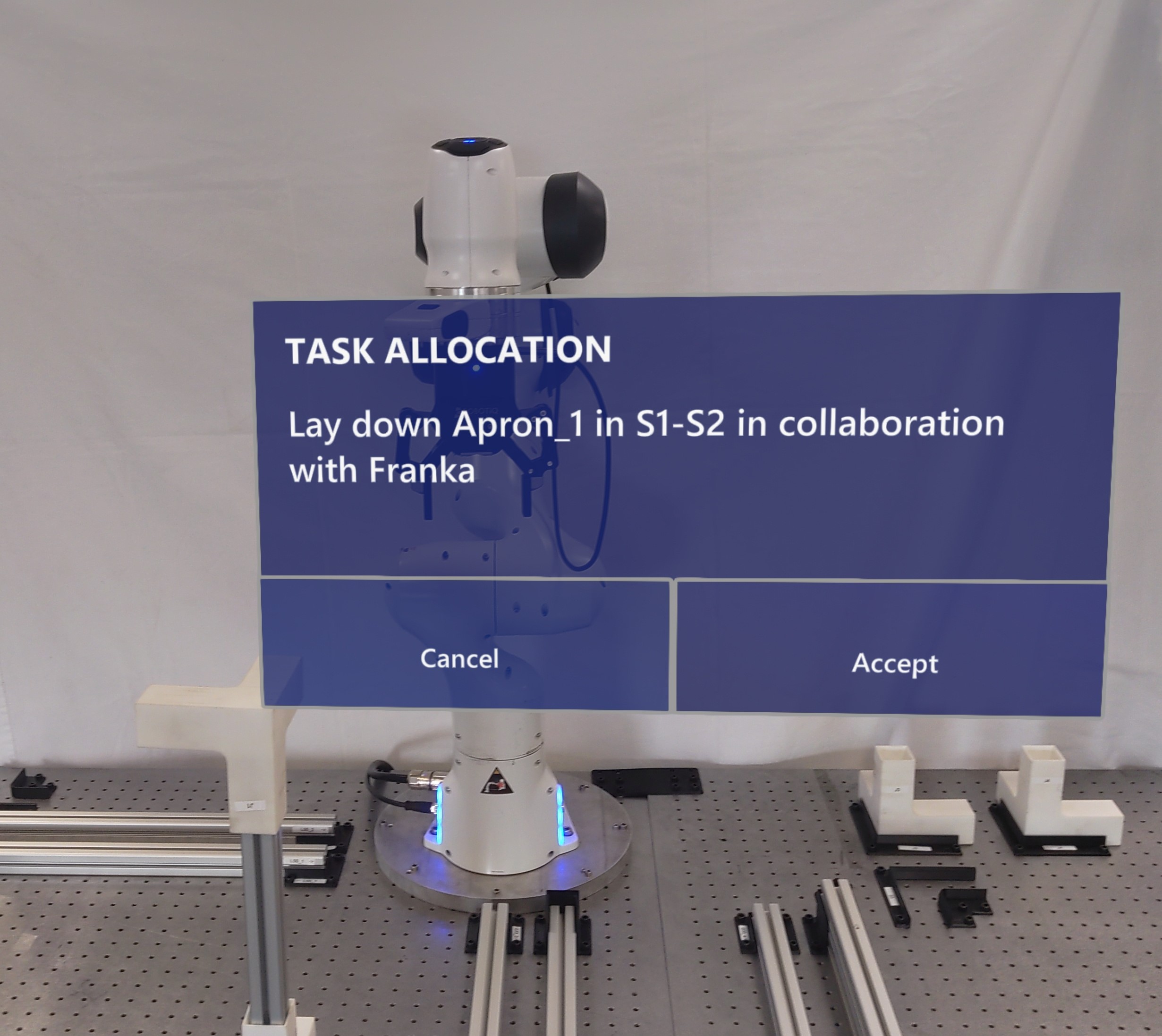}
	     \caption{Collab. allocation.}
	     \label{fig:UI_collab_alloc}
     \end{subfigure}
     \hfill
     \begin{subfigure}[b]{0.19\textwidth}
         \centering
    	 \includegraphics[trim=0.0cm 0.0cm 0.0cm 14.5cm,clip,width=3.2cm,height=2cm]{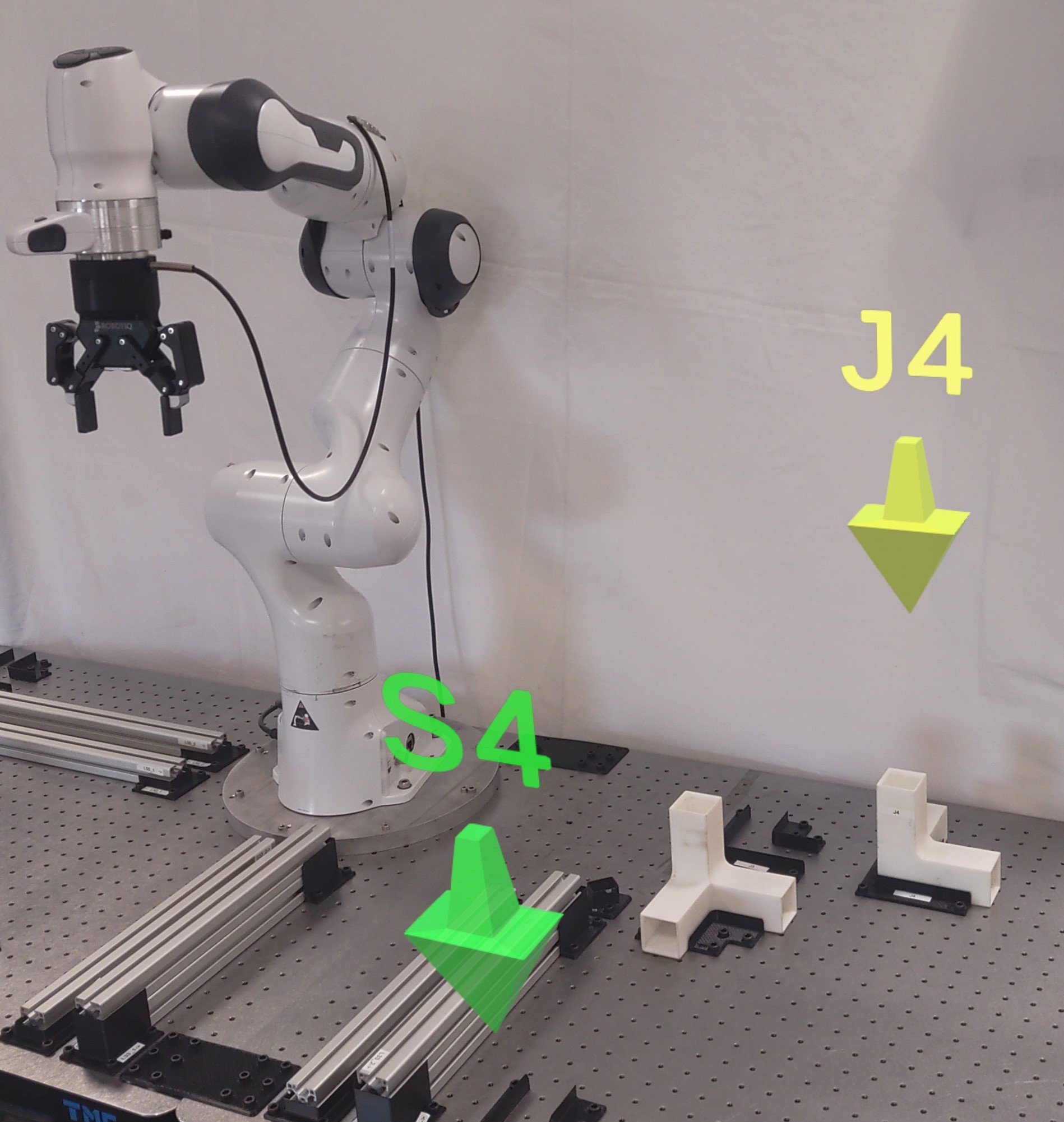}
	     \caption{Piece insertion.}
	     \label{fig:UI_instructions_1}
     \end{subfigure}
     \hfill
     \begin{subfigure}[b]{0.19\textwidth}
         \centering
	     \includegraphics[trim=0.0cm 0.0cm 0.0cm 0.0cm,clip,width=3.2cm,height=2cm]{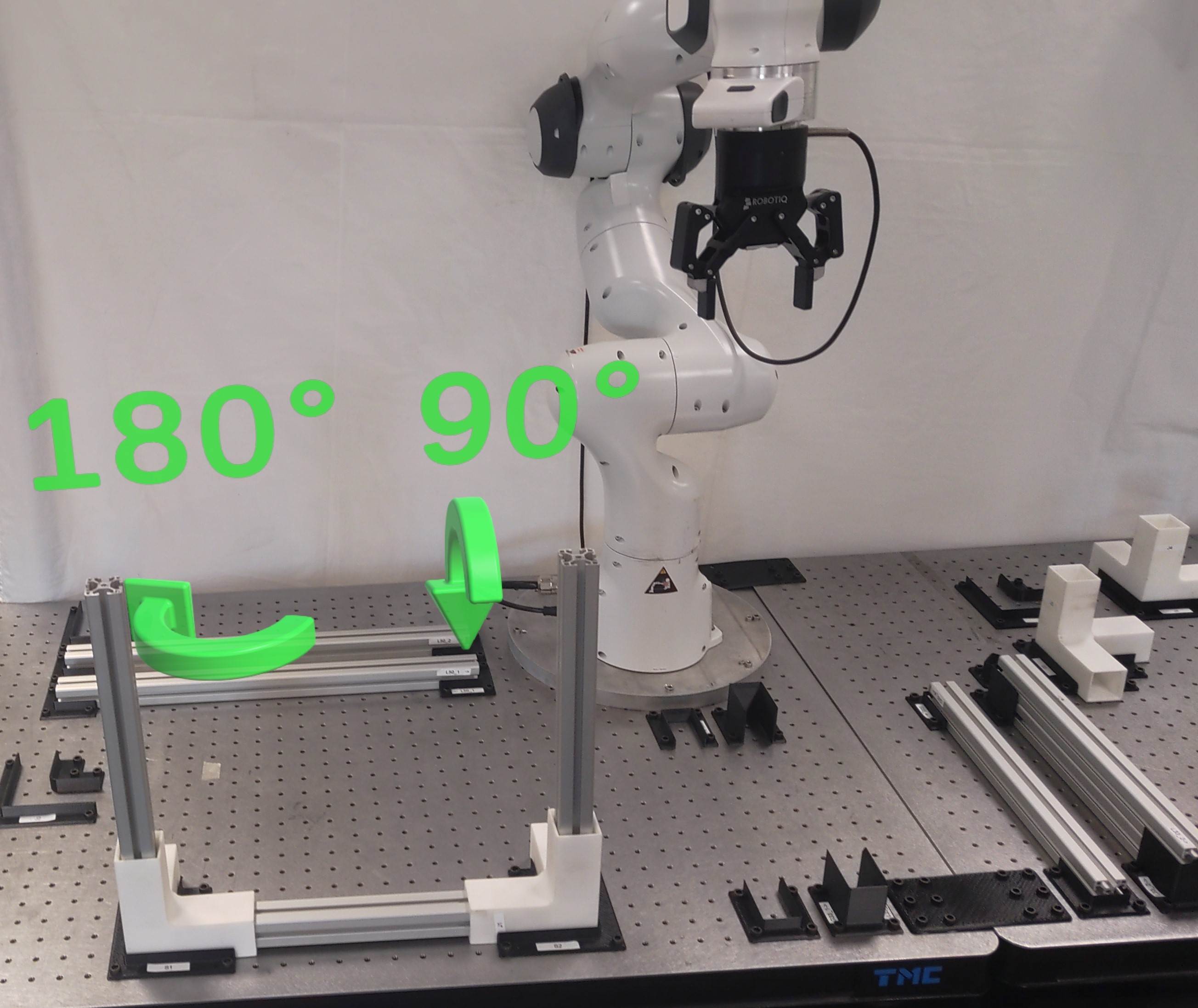}
	     \caption{Piece rotation.}
	     \label{fig:UI_instructions_2}
     \end{subfigure}
     \hfill
     \begin{subfigure}[b]{0.19\textwidth}
        \centering
	\includegraphics[trim=0cm 0.0cm 0cm 0cm,clip,width=3.2cm,height=2cm]{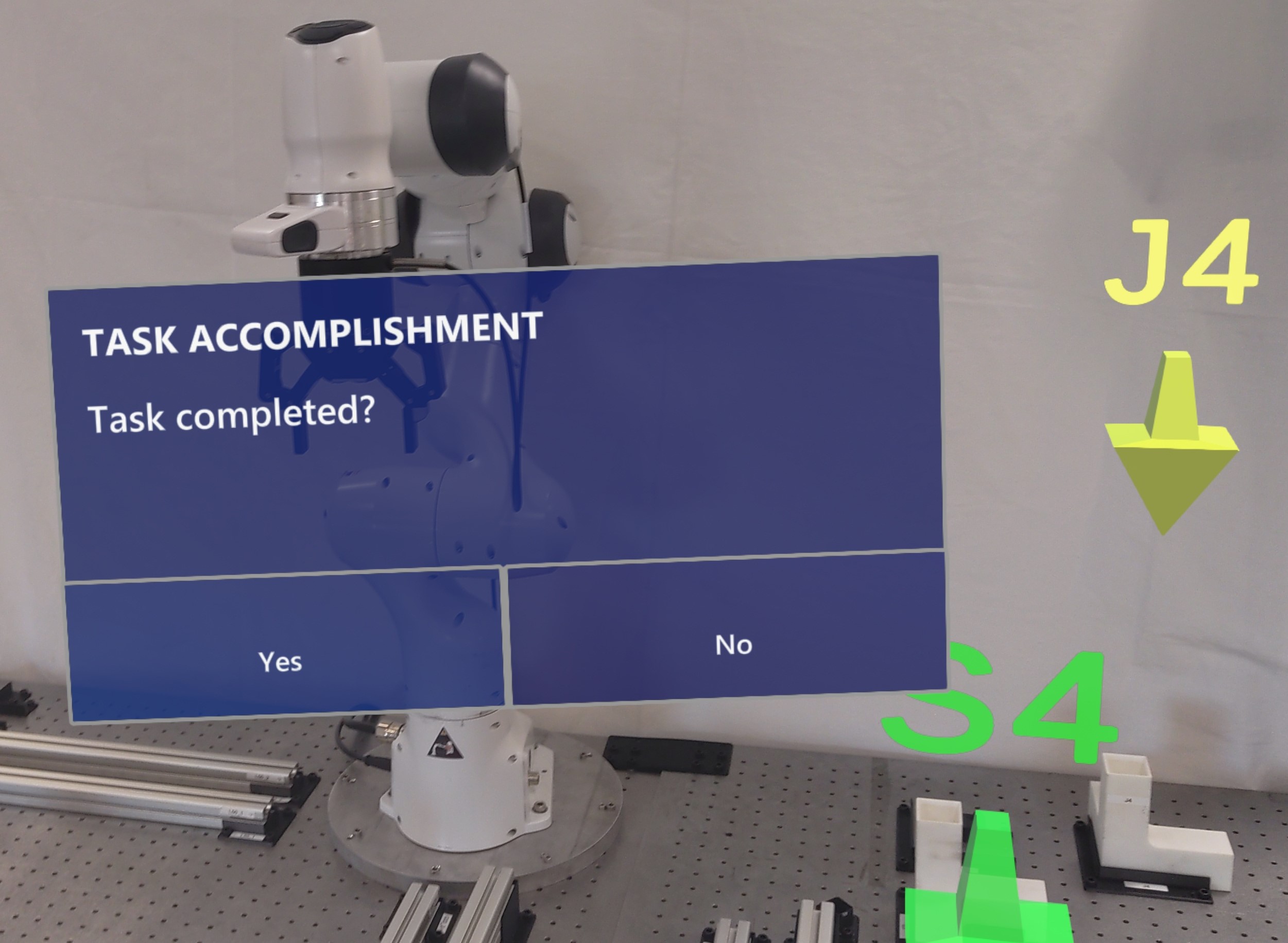}
        \caption{Action Completed.}
        \label{fig:UI_completed}
     \end{subfigure}
     \caption{Snapshots of the holograms included in the AR-based User Interface.}
     \label{fig:UI_pics}
\end{figure*}

To model the suitability of the workers $w_{a\Join b}$ for a single action $a_j$, in a collaborative performance, two directions can be taken. The simplest one consists in computing the capability indicator $c_{(a\Join b )j}$ as a function of the individual worker capabilities for the same action, namely $c_{(a\Join b )j} = f(c_{aj},c_{bj})$. The main advantage of this solution consists in the fact that the computation of the $M \times P$ costs boils down to the computation of $M \times N$ workers capabilities. On the other hand, modeling the suitability in the collaborative scenario as a function of the individual worker suitability is not simple, and there's no guarantee that the individual suitability is descriptive of the teamed performance, that, moreover, depends on the specific design of the collaborative task.
For these reasons, we opted to capitalize on capability indicators that are explanatory of both single worker performance and also the teamed performance. A practical example of such costs is explained in \autoref{sssec:exp_metrics}.

\paragraph{Designing the Preference Cost}
\label{par:preference_cost}
This cost is added to introduce the preference of a human worker in performing a specific action. In particular, it is defined as:
\begin{equation}
    \psi_{ij} = \Psi_i \frac{\#\;negations_{ij}}{\#\;negotiations_{ij}}
\end{equation}
where $\Psi_i$ is a constant gain, while $\#$ $negations_{ij}$ and $\#$ $negotiations_{ij}$ are set in the negotiation phase of worker $w_i$ for action $a_j$. If $i=a\Join  b$, i.e., a collaboration between $w_a$ and $w_b$, where one of the two workers is human, the collaboration preference is updated.   


\paragraph{Designing the Availability Costs}
\label{par:availability_cost}
We consider a worker available if it is present in the workcell and if it is not occupied by any other action. In case multiple workers are available, the allocator node tries to assign the action to the worker whose cost is smaller.
The availability value has been added to the cost function, and not to the problem constraints, since we cannot ensure that at least an worker is always available.
One simple definition of availability cost is the following:
\begin{equation} \label{eq:binary_availability}
    \xi_i = \begin{cases}
    0, \quad &\text{if } w_i \text{ is } \textit{available};  \\
    \alpha_i, \quad &\textit{otherwise.}
    \end{cases}
\end{equation}
$\alpha_i$ is the unavailability cost of $w_i$. To ensure that the availability weights more than the other costs, we set $\alpha_i > \max{\{c_{ij}\}}$. With this method, we are able to ensure that the allocation node favors an available worker, minimizing in this way the single-worker waiting times. In this case, the worker's suitability for the action is not considered. 
On the other hand, in some situations, might be convenient to make the system wait for a suitable worker (low Worker-Action cost), instead of assigning the action to one available agent. This is the case, for example, when other workers are available, but are all unsuitable for the action (high Worker-Action cost).
For this reason, we modified the binary nature of the availability to account for the remaining execution time:
\begin{equation} \label{eq:availability}
    \xi_i = \begin{cases}
    0, \quad &\text{if } w_i \text{ is } \textit{available};  \\
    \alpha_i \frac{T_{a_{ij}} - t_{a_{ij}}}{T_{a_{ij}}} , \quad &\textit{otherwise.}
    \end{cases}
\end{equation}
where $T_{a_{ij}}$ is the nominal duration of $a_j$ performed by $w_i$, $t_{a_{ij}}$ is the time spent by $w_i$ for $a_j$ from the beginning of $a_j$, where $t_{a_{ij}} \le T_{a_{ij}}$. In this way, $\frac{T_{a_{ij}} - t_{a_{ij}}}{T_{a_{ij}}}=1$ if the action has just started, and, as the worker completes the action, it goes to 0.  
To ensure that costs are comparable, $\alpha_i = \Psi_i = \max{\{c_{ij}\}}$, $\forall\; j$.

%% file: sections/user_interface.tex
To enable two-way communication with the allocation architecture, human workers are provided with a user interface (UI), that allows the operators not only to receive instructions and visual feedback but also to acknowledge the system about their decisions. 
The interface capitalizes on an Augmented Reality (AR) headset. Through the creation of a real-world anchor, in which a virtual object is placed in a fixed pose with respect to the physical world, the app is able to identify the workstation and, therefore, to place action instructions and holograms in specific positions of the real world. Hence, the AR interface enables the human co-workers to perceive the holograms superimposed to real world objects. The AR app is constantly in communication with the system for data exchange through the network, including service requests or messages broadcast.

Once the app is launched, it calibrates the AR device in order to detect the environment exploiting the identification of the real-world anchor. After the workstation is recognized, the pose of the anchor with respect to the camera of the AR device is computed. In this way, such 3D coordinates are used to spawn accordingly to the real environment. 
When the user logs-in, the system sends a request to each human worker with the allocated action (\autoref{fig:UI_alloc}) by a written instruction in a dialog window.
The operator is also informed if the action should be accomplished in collaboration with another worker (\autoref{fig:UI_collab_alloc}). The window is composed by two buttons: \textit{Accept} and \textit{Cancel}. Hence, the worker can decide to accept or reject the action just by clicking on the corresponding button. If an action is rejected, the system is advised and, after re-computing the allocation for that specific worker, sends another request with the new allocated action. Instead, if an action is accepted, two possible types of virtual instructions can appear in the environment: either yellow and green vertical arrows or green rounded arrows. If the action requires a pick and place or an insertion, the instruction is represented by two holograms represented by vertical arrows (\autoref{fig:UI_instructions_1}), each one with the name of the object or the station on top. The yellow arrow points the piece that has to be moved or inserted, while the green one is target location or object for the piece. If instead a rotation of the object is needed, the virtual instructions are composed of one or more circular green arrows with the angle on top, representing the direction of the rotation and the angle degrees, respectively (\autoref{fig:UI_instructions_2}). If the circular arrows are more than one, the instructions should be followed from the left to the right. In order to inform the system of the completion of the action, the user has to perform a gesture in the field-of-view of the device, i.e. an \textit{air-tap} or touch simulation. Once the gesture is detected, a dialog window with two buttons, i.e. \textit{Yes} and \textit{No}, appears asking if the action is completed (\autoref{fig:UI_completed}). While the worker is performing an action, a fault \textit{air-tap} could be detected making the windows show up. For this reason, the user has the ability to click on the \textit{No} button. If this is the case, the app is waiting for another \textit{air-tap} gesture for the action completed communication.

%% file: sections/experiments.tex
The performance of the proposed approach is evaluated through several simulation and real-world experiments. The simulation experiments were conducted to initially characterize the method in terms of computational complexity and of allocation results in different conditions, to compare the results with other approaches present in the literature. The real-world experiments focuses on the allocation results with single index costs (performance and ergonomics) with an usability analysis of the framework with a subjective evaluation in a multi-subject experiment. In particular, the algorithms compared are the following:
\begin{itemize}
    \item \textit{Collaborative Multi-Task Behavior Tree (Collab-MT-BT)}: this algorithm features the custom Behavior Trees presented in \autoref{ssec:hrc_model} as task model and the allocation method that includes also collaborative action execution, explained in \autoref{sssec:collab_model}.  
    \item \textit{Cooperative Multi-Task Behavior Tree (Coop-MT-BT)}: this algorithm presents the same task structure as in Collab-MT-BT, but no collaborative actions can be planned. The job is modeled as in \autoref{ssec:hrc_model} and can allocate actions only to single workers according to \autoref{sssec:coop_model}.
    \item \textit{Cooperative Single-Task Behavior Tree (Coop-ST-BT)}: this algorithm exploits both job model and allocation method in \cite{fusaro2021integrated}. Actions are planned only cooperatively.
    \item \textit{AND/OR graph}: this algorithm exploits AND/OR graphs to model the job and an AO* to solve the allocation problem with respect to single workers~\cite{johannsmeier2017hierarchical,darvish2018flexible}. Also, this method does not envision collaborative action execution. The details of the implementation of this method can be found in~\cite{merlo2022dynamic}.
\end{itemize}

The framework was tested on a laptop with an Intel Core i7-8565U 1.8 GHz $\times$ 8-cores CPU and 8 GB RAM. The architecture has been developed in C++, on Ubuntu 18.04 and ROS Melodic, exploiting the \href{https://github.com/BehaviorTree/BehaviorTree.CPP}{BehaviorTree.CPP} library to define the BT nodes and the \href{https://github.com/coin-or/Osi}{Osi} library with the GNU Linear Programming Kit solver to formalize the MATA problem. The user interface, used in the real-world experiments, is developed in \href{https://unity.com/}{Unity} which allows to deploy the application in different AR platforms. The anchor of a virtual object in the real world environment is implemented using a Unity package provided by \href{https://github.com/Azure/azure-spatial-anchors-samples}{Microsoft Azure}, while the holographic representation is created through the \href{https://github.com/microsoft/MixedRealityToolkit-Unity}{Microsoft Mixed Reality Toolkit}. The communication layer between ROS and Unity is built exploiting the \href{https://github.com/Unity-Technologies/ROS-TCP-Endpoint}{ROS-TCP-Endpoint} package. Moreover, the ROS environment and the Unity application deployed on the AR device are connected to the same local network to ensure data exchange.

\subsection{Computational Complexity and System Scalability} \label{ssec:complexity}
\input{sections/computational_complexity}

\subsection{Simulations} \label{ssec:simulations}
\input{sections/simulations}

\subsection{Real-World Experiments} \label{ssec:real_exp}
\input{sections/real_exp}

%% file: sections/computational_complexity.tex
\begin{figure}[t]
    \centering
    \includegraphics[trim=3.8cm 0.8cm 4.5cm 1cm,clip,width=0.8\linewidth]{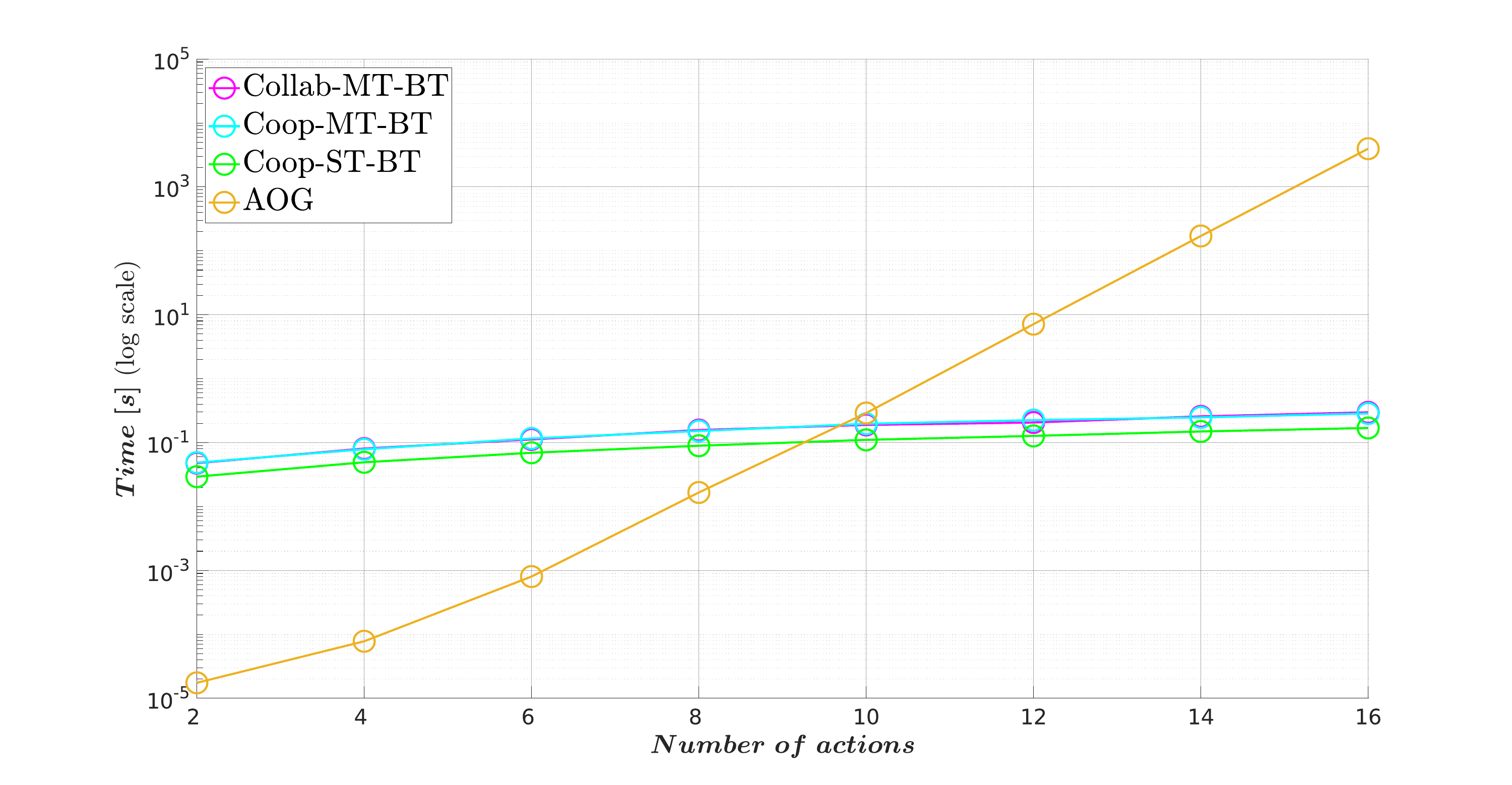}
    \caption{Computational time (in log scale) with 3 workers and increasing number of actions. The trend of the Collab-MT-BT (pink) matches the Coop-MT-BT (cyan).}
    \label{fig:complex_act}
    \vspace{-2mm}
\end{figure}

\begin{figure}[t]
    \centering
    \includegraphics[trim=5cm 1.2cm 4.5cm 2cm,clip,width=0.8\linewidth]{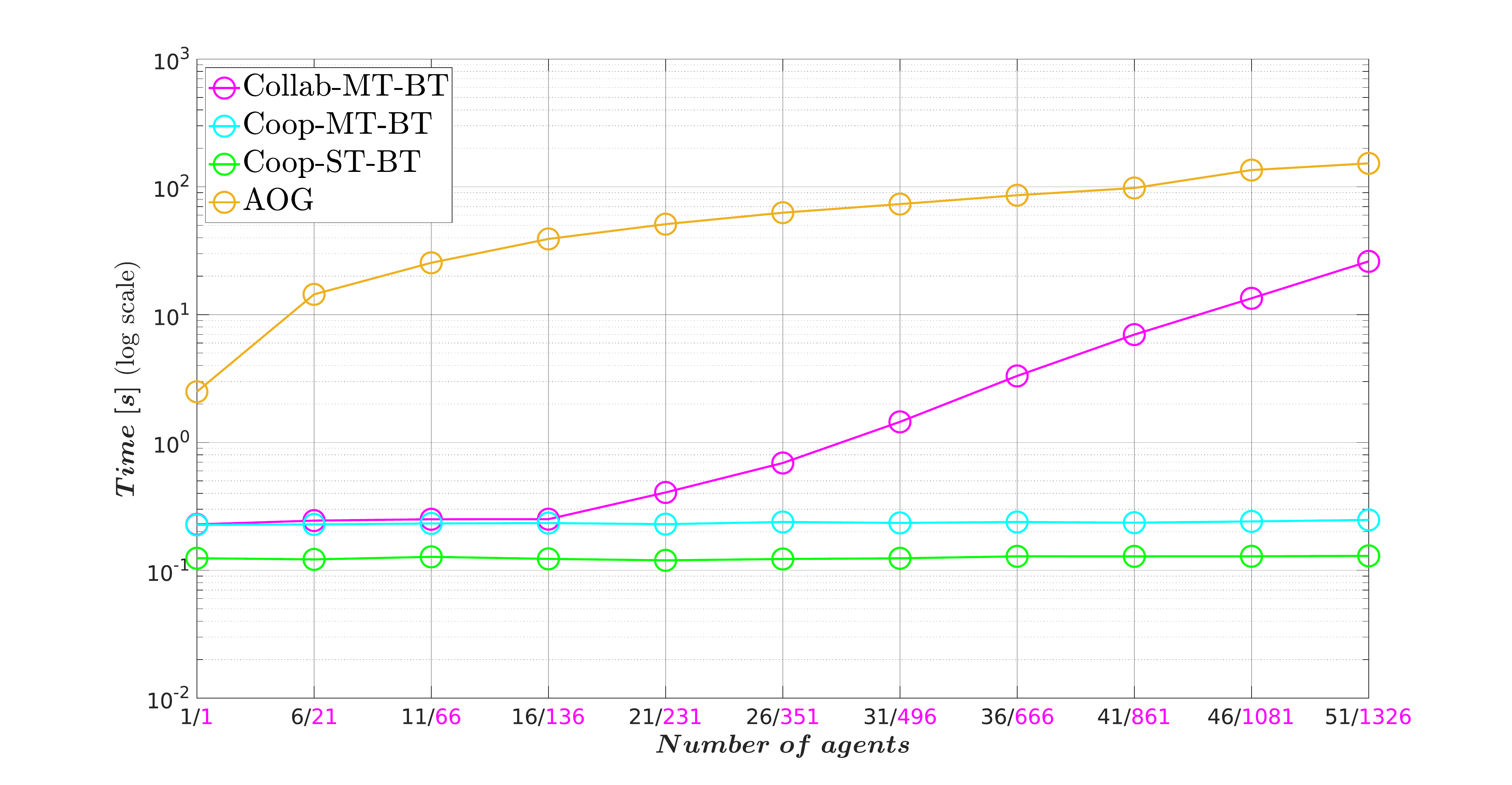}
    \caption{Computational time (in log scale) with 12 actions and increasing number of workers. In the x-axis are reported the number of total agents (black) and the number of candidates (pink) of the allocation in case of a collaborative scenario as in \eqref{eq:W_workers}. In the cooperative scenario, the the number of total agents and the number of candidates coincides. }
    \label{fig:complex_agt}
    \vspace{-2mm}
\end{figure}

First, the computational complexity of the allocation algorithm is assessed. The computation time is defined as the time from the initialization of the algorithm until the action execution, excluding the execution time of the action, averaged on 10 repetitions of the same BT.
To obtain a comprehensive evaluation and showcase the performances in a wide range of possible scenarios, the size of the problem is progressively expanded by increasing the number of actions and the number of workers, separately.

We compared the time complexity of our approach with the gold standard of graph-based role allocation methods for assembly tasks, the \textit{AND/OR graph}~\cite{johannsmeier2017hierarchical,darvish2018flexible}, to be used as a baseline, as in~\cite{darvish2021hierarchical} and the one presented by the authors in a previous work~\cite{fusaro2021integrated}, which still exploits BTs as task planners but does not foresee by definition any collaborative action allocation (and hence execution) and does not allow a complete parallelization of tasks.
Since AND/OR graphs also model assembly tasks where all actions are in sequence, we compared the performances with tasks composed of actions only in sequence. It is important to highlight that our method can model a more general class of industrial tasks with respect to assemblies and is not limited to sequential plans.

\begin{figure}[t]
    \centering
    \includegraphics[trim=3.8cm 0.8cm 4.1cm 1.5cm,clip,width=0.8\linewidth]{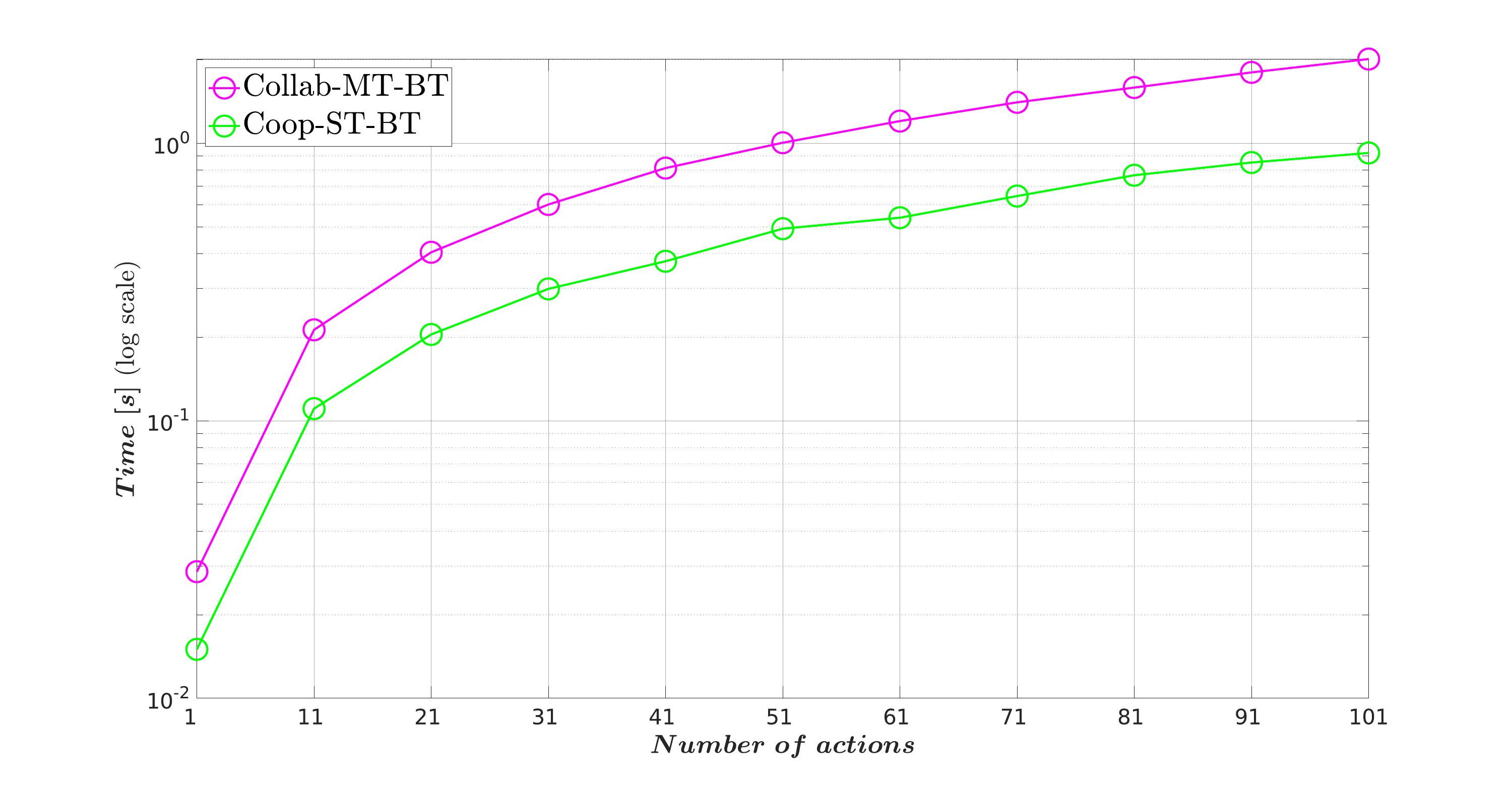}
    \caption{Computational time (in log scale) with 4 workers and increasing number of actions in series.}
    \label{fig:complex_old_new_series}
    \vspace{1mm}
\end{figure}

\begin{figure}[t]
    \centering
    \includegraphics[trim=3.8cm 0.8cm 4.1cm 1.5cm,clip,width=0.8\linewidth]{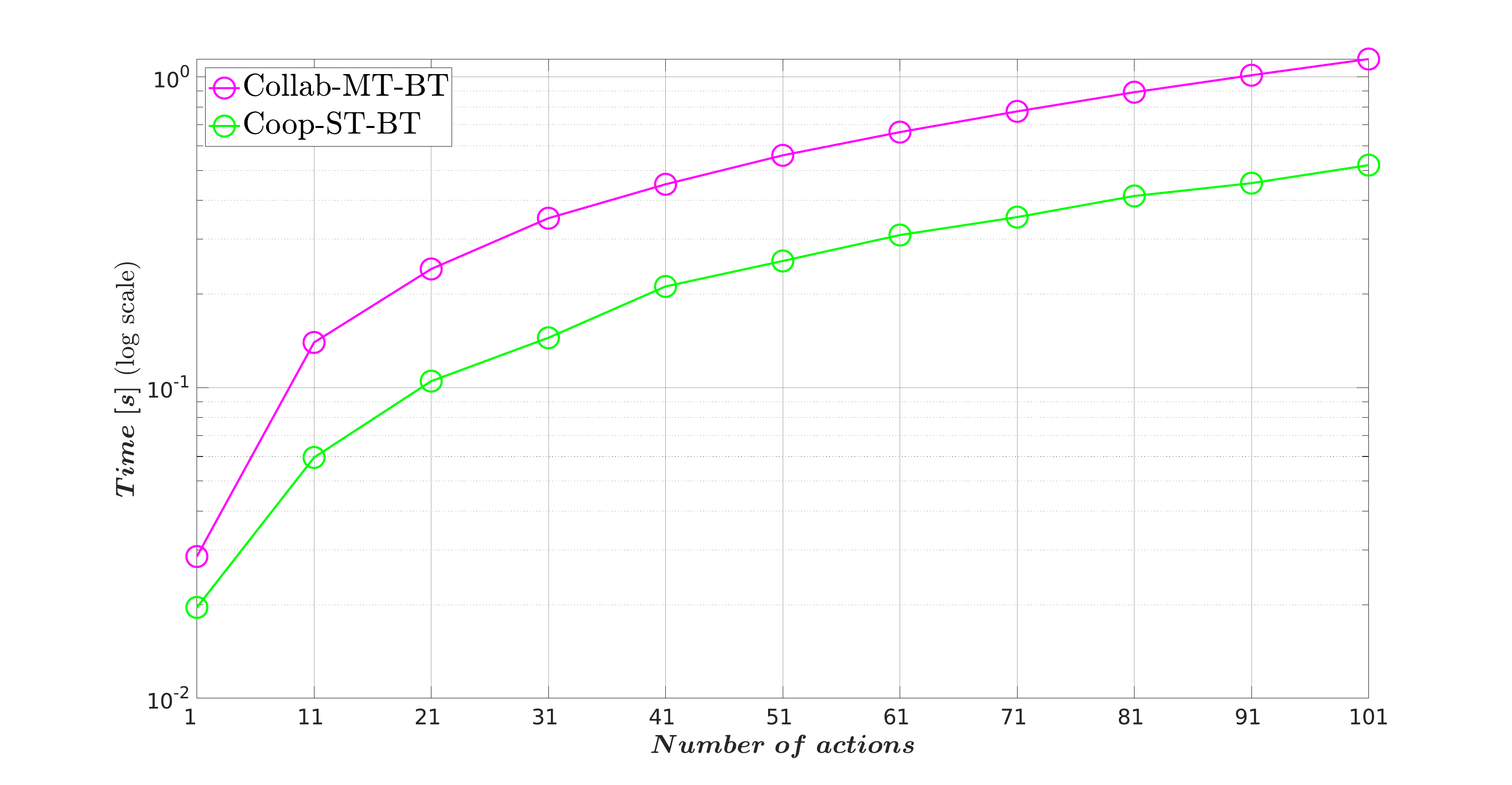}
    \caption{Computational time (in log scale) with 4 workers and increasing number of actions in parallel.}
    \label{fig:complex_old_new_parallel}
    \vspace{-2mm}
\end{figure}

In the plots in \autoref{fig:complex_act} the number of workers is fixed to 3 (small-sized team) and the number of actions is increased from 2 to 16. It is interesting to see that, as the number of actions increases with a reasonable task size ($>10$), AND/OR graphs are outperformed by the BT-based algorithms.
The maximum number of actions in this comparison was limited by the performance of the AND/OR graph, which took approx. 24 hours to solve the role allocation problem with 3 workers with more than 16 actions.  
The computational complexity was also evaluated by increasing the number of workers while keeping the number of actions fixed at 12 (due to the limitations of AND/OR graphs performances). The comparison is illustrated in \autoref{fig:complex_agt}, in which it can be observed that the performance of the BT methods is computationally less expensive. The \textit{Collab-MT-BT} highlights an increase in the computational time due to the possible collaborations between the workers, represented by the pink number in the x-axis, which is higher than the number of workers, as explained in \eqref{eq:W_workers}.
However, with a reasonable number of workers for a single job (e.g., $<21$), the performance of the BTs algorithms is similar and way less expensive with respect to AND/OR graphs.

To further evaluate the method beyond the limitations imposed by the performance of the AND/OR graphs, the \textit{Collab-MT-BT} and \textit{Coop-ST-BT} methods were compared, increasing the number of actions from 1 to 101, keeping the number of workers fixed at 4 (hence, for \textit{Collab-MT-BT} the number of workers includes the collaborations, for a total of 10 potential candidates). It can be seen that in both cases, for all actions performed in series (\autoref{fig:complex_old_new_series}) and in parallel (\autoref{fig:complex_old_new_parallel}), the complexity values and trends are similar for both approaches.

\begin{table*}[t]
\begin{center}
\begin{tabular}{|c||c|c|c|c|c|c|c|c|c|}
\hline
\multicolumn{1}{|c||}{\multirow{2}{*}{\textbf{Action}}} & \multirow{2}{*}{$\boldsymbol{w_1} (s)$} & \multirow{2}{*}{$\boldsymbol{w_2} (s)$} & \multirow{2}{*}{$\boldsymbol{w_3} (s)$} & \multicolumn{1}{c|}{$\boldsymbol{w_1\Join w_2} (s)$} & \multicolumn{1}{c|}{$\boldsymbol{w_2\Join w_3} (s)$} & \multicolumn{1}{c|}{$\boldsymbol{w_1\Join w_3} (s)$} & \multicolumn{3}{c|}{\textbf{Worker allocated}} \\[0.7mm] \cline{5-6} \cline{7-8} \cline{9-10}
\multicolumn{1}{|c||}{} & & & & \textbf{Collab-MT} & \textbf{Collab-MT} & \textbf{Collab-MT} & \inlinebox[colback=yellow,colframe=electricpink]{\textbf{Collab-MT}} & \inlinebox[colback=yellow,colframe=darkblue]{\textbf{Coop-MT}} & \inlinebox[colback=yellow,colframe=bluette]{\textbf{Coop-ST}}\\[0.7mm] \hhline{----------}
$a_1$ & 15 & 20 & 25 & 32 & 33 & 29 & \cellcolor{blue!50} $w_1$ & \cellcolor{blue!50} $w_1$ & \cellcolor{blue!50} $w_1$\\[0.7mm]
$a_2$ & 27 & 22 & 20 & 33 & 31 & 32 & \cellcolor{green!50} $w_3$ & \cellcolor{green!50} $w_3$ & \cellcolor{green!50} $w_3$\\[0.7mm]
$a_3$ & 17 & 21 & 19 & 25 & 27 & 12 & \cellcolor{olive!50} $w_1\Join w_3$ & \cellcolor{blue!50} $w_1$ & \cellcolor{blue!50} $w_1$\\[0.7mm]
$a_4$ & 13 & 14 & 11 & 9 & 20 & 17 & \cellcolor{purple!50} $w_1\Join w_2$ & \cellcolor{blue!50} $w_1$ & \cellcolor{green!50} $w_3$\\[0.7mm]
$a_5$ & 18 & 17 & 25 & 27 & 32 & 30 & \cellcolor{orange!50} $w_2$ & \cellcolor{orange!50} $w_2$ & \cellcolor{orange!50} $w_2$\\[0.7mm]
$a_6$ & 27 & 29 & 31 & 36 & 40 & 38 & \cellcolor{blue!50} $w_1$ & \cellcolor{blue!50} $w_1$ & \cellcolor{blue!50} $w_1$\\[0.7mm]
$a_7$ & 37 & 35 & 27 & 41 & 47 & 42 & \cellcolor{green!50} $w_3$ & \cellcolor{green!50} $w_3$ & \cellcolor{green!50} $w_3$\\[0.7mm]
$a_8$ & 38 & 33 & 39 & 45 & 43 & 44 & \cellcolor{orange!50} $w_2$ & \cellcolor{orange!50} $w_2$ & \cellcolor{orange!50} $w_2$\\[0.7mm]
$a_9$ & 27 & 25 & 24 & 30 & 34 & 31 & \cellcolor{green!50} $w_3$ & \cellcolor{green!50} $w_3$ & \cellcolor{green!50} $w_3$\\[0.7mm]
$a_{10}$ & 13 & 19 & 18 & 11 & 25 & 23 & \cellcolor{purple!50} $w_1\Join w_2$ & \cellcolor{blue!50} $w_1$ & \cellcolor{blue!50} $w_1$\\[0.7mm]
$a_{11}$ & 17 & 12 & 20 & 15 & 23 & 24 & \cellcolor{orange!50} $w_2$ & \cellcolor{orange!50} $w_2$ & \cellcolor{orange!50} $w_2$\\[0.7mm]
$a_{12}$ & 31 & 25 & 24 & 38 & 37 & 36 & \cellcolor{green!50} $w_3$ & \cellcolor{green!50} $w_3$ & \cellcolor{green!50} $w_3$\\[0.7mm]
$a_{13}$ & 10 & 9 & 12 & 15 & 7 & 18 & \cellcolor{pink!50} $w_2\Join w_3$ & \cellcolor{orange!50} $w_2$ & \cellcolor{orange!50} $w_2$\\[0.7mm]
\hline
\end{tabular}
\end{center}
\caption{Allocation results of the actions of the simulated job and corresponding actions costs in the different scenarios. Each color corresponds to a different allocated worker.}
\label{table:allocation_results}
\vspace{-2mm}
\end{table*}

%% file: sections/simulations.tex
\begin{figure}[t]
    \centering
    \includegraphics[trim=0cm 0.5cm 0cm 0cm,clip,width=0.8\linewidth]{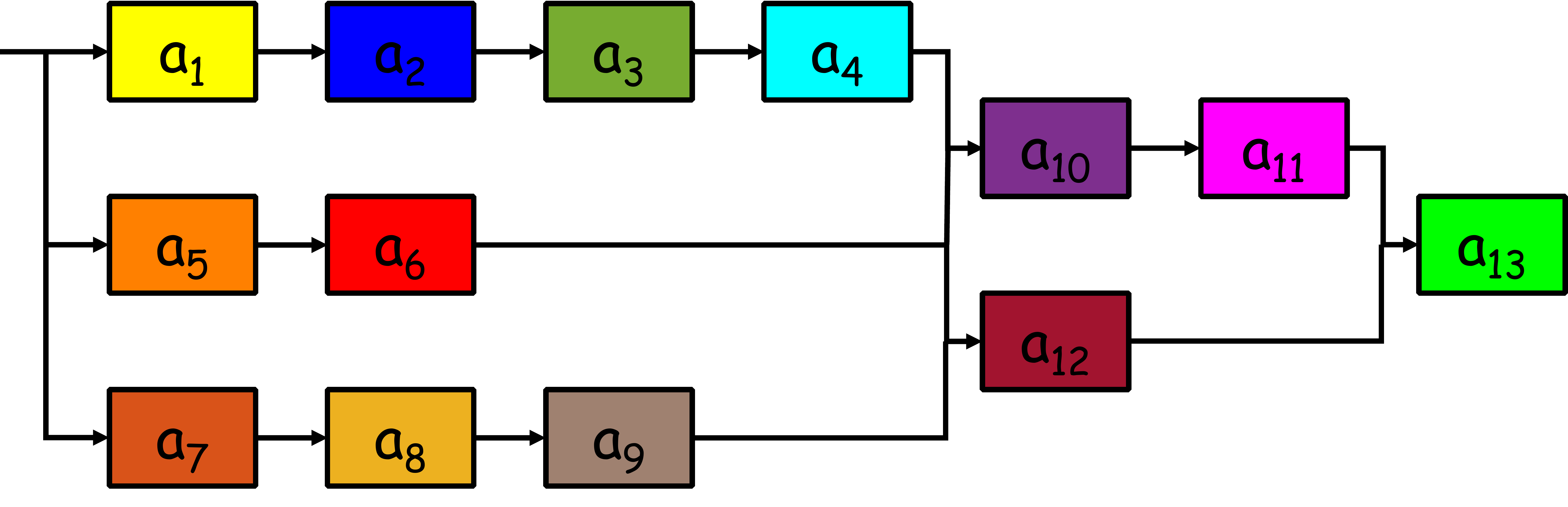}
    \caption{Task plan of the simulated job.}
    \label{fig:sim_job_seq}
    \vspace{-2mm}
\end{figure}

\begin{figure}[t]
    \centering
    \includegraphics[trim=0.2cm 0.2cm 0cm 0.5cm,clip,width=0.95\linewidth]{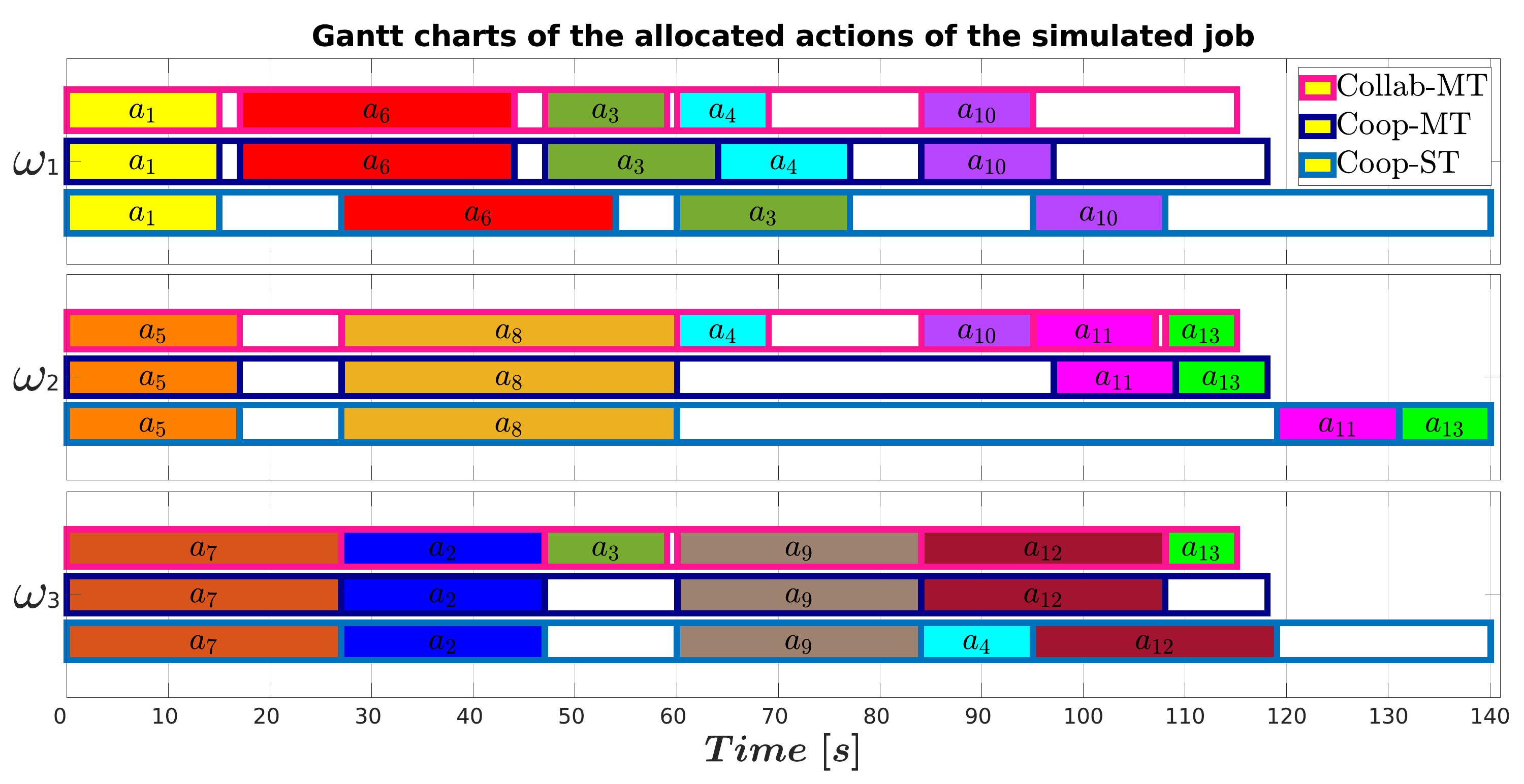}
    \caption{Gantt charts of the simulated job with allocated actions per worker with the three allocation methods, depicted with different colored outlines. White rectangles represent times when the worker is inactive. The colors of the actions refer to the action scheme in \autoref{fig:sim_job_seq}.}
    \label{fig:gantt_chart}
    \vspace{-2mm}
\end{figure}

To evaluate the planning and allocation capabilities of the new method with respect to the approach in~\cite{fusaro2021integrated} a simulation was performed with a target job composed of 13 actions performed by 3 workers. The job is fictional, and the task structure is handcrafted by the authors. For this reason, actions are not specified in this experiment.
We assume that each worker executes the assigned action within the specified execution time.
The scheme of the complete job, and hence the temporal and logic constraints of the actions, is shown in \autoref{fig:sim_job_seq}. In this experiment, the considered collaborations are limited to pairs, as in \eqref{eq:W_agents_2}, and the costs are defined as the execution time of each action (in seconds) for each worker or pair of workers (in \textit{Collab-MT}\footnote{From now on, the -BT suffix will be dropped, as the compared methods include the BT as task planner.}), as in \autoref{table:allocation_results}. From the \autoref{fig:gantt_chart}, it can be noticed that in the \textit{MT} methods, not only is there a reduction of the overall execution time but also of the waiting times of each worker. This is because the \textit{MT} methods overcome the limitations on the task structure imposed by the \textit{ST} approach, enabling the parallelization of sequences. 
For example, in this experiment, as shown in \autoref{fig:sim_job_seq}, $a_1$, $a_5$, and $a_7$ can be performed at the same time. Once one of the actions is completed, e.g. $a_1$, the next one should be immediately allocated, e.g. $a_2$, but, due to its structure, the \textit{ST} method has to wait for all three actions in order to allocate the next ones. Moreover, it can also be noticed, given the costs in~\autoref{table:allocation_results}, that there is a reduction in the \textit{Collab-MT} overall execution time compared to \textit{Coop-MT}. 
This is possible if the collaboration is faster than each single agent's execution. Otherwise, \textit{Collab-MT} will output the same solution as the \textit{Coop-MT}, as the latter's formulation is included in the former.

%% file: sections/real_exp.tex
The \textit{Collab-MT-BT} strategy was also evaluated with a collaborative assembly job in a flexible production line scenario, consisting of a proof-of-concept table assembly. With this setup, three main experiments are performed: i) the evaluation of the effectiveness of the architecture with AR interface with respect to the same action executed manually without AR interface (\autoref{sssec:exp_hrc_ar}) ; ii) a comparison of the allocation results using performance-based and human stress-based metrics (\autoref{sssec:exp_metrics}); and iii) usability testing through a multi-subject evaluation (\autoref{sssec:exp_user_study}). Note that the system was tested with three modes (H, H+AR, HRC+AR) by 25 subjects, while HRC+AR with a stress-based strategy was tested by only one of the 25 subjects. In particular, in the first two experiments, the results compare only one execution.

\subsubsection{Experimental Setup} \label{sssec:exp_setup}
\begin{figure}
    \centering
    \includegraphics[trim=2cm 1cm 1cm 0cm,clip,width=0.6666\linewidth]{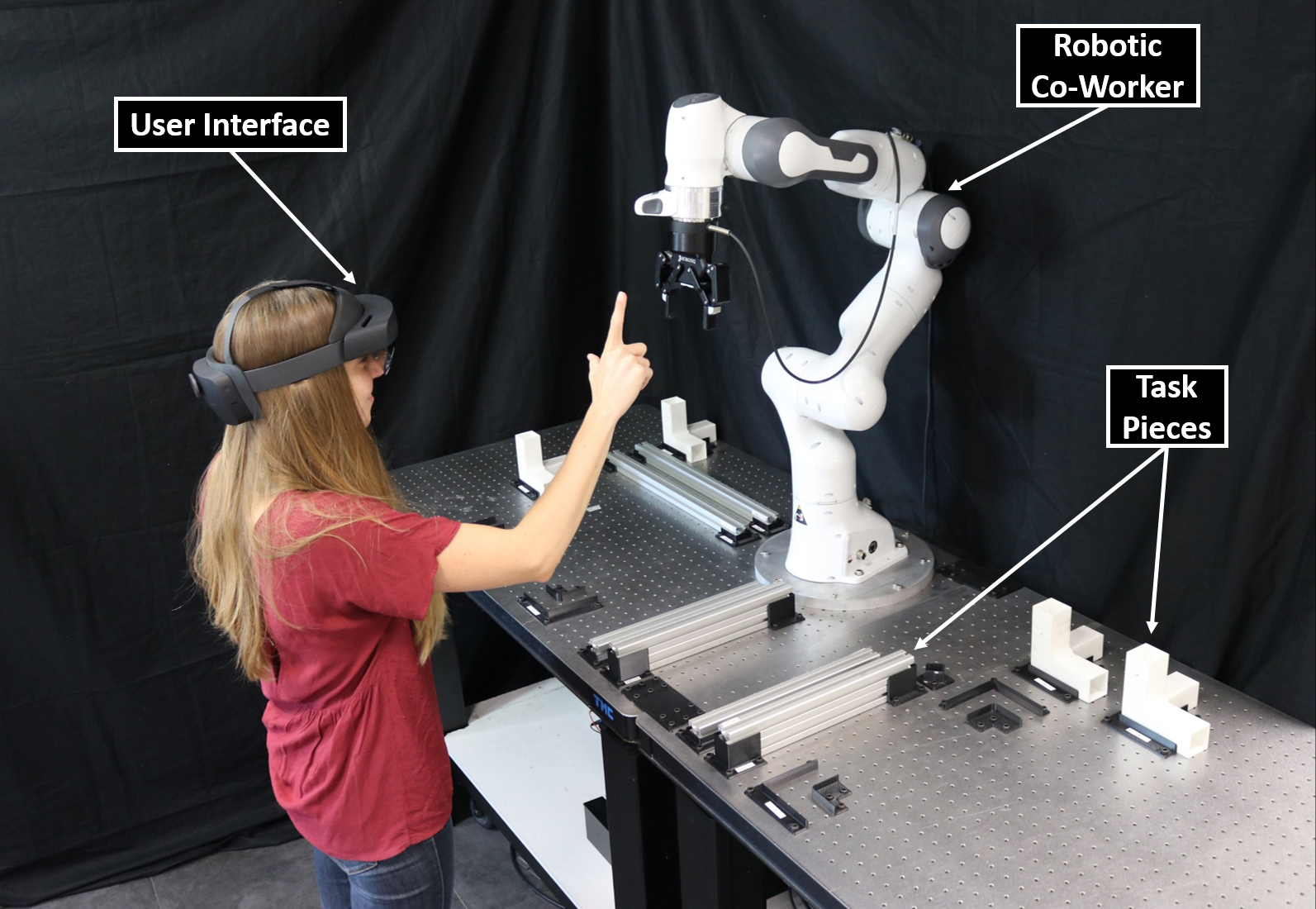}
    \caption{Experimental setup. The experiments are performed by two workers, a human equipped by an AR user interface and a fixed manipulator. The task pieces and the stations in which the assembly is performed, are placed in specific positions on the tables.}
    \label{fig:exp_setup}
    \vspace{-2mm}
\end{figure}

\begin{figure}
    \centering
    \includegraphics[trim=0cm 0cm 0cm 0cm,clip,width=0.9\linewidth]{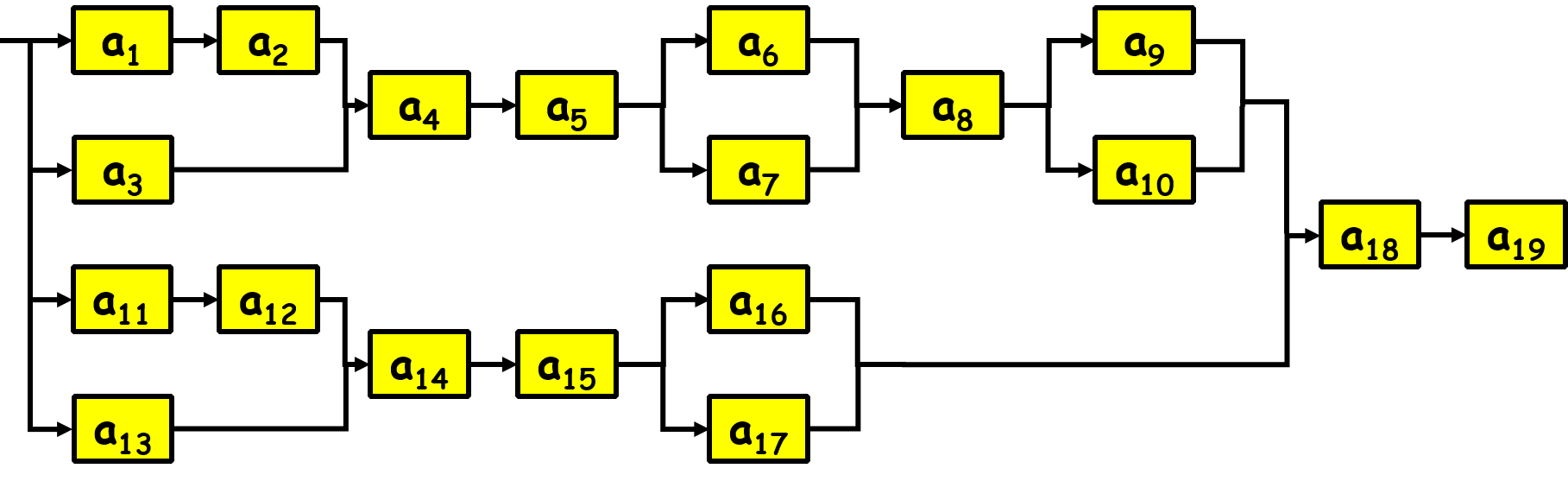}
    \caption{Assembly task plan of the table assembly job. The plan can be divided in two main parallel sequences in which the two aprons of the table are assembled (corresponding to the two tables of the setup in \autoref{fig:exp_setup}). Then, the two aprons are assembled together in the last two actions.}
    \label{fig:real_job_seq}
    \vspace{-2mm}
\end{figure}

The table was built with aluminum profiles of different sizes, held together magnetically through 3D printed joints. 
The whole assembly job is made up of 19 actions, whose temporal and logic constraints are shown in \autoref{fig:real_job_seq}. 
In collaborative action executions, the team is composed of a Franka Emika Panda manipulator equipped with a two-finger gripper Robotiq 2F-85 and a human worker equipped with the Microsoft HoloLens 2 headset running the AR user interface. Two clutched tables, on which the pieces and assembly stations were placed, represented the shared workspace between the two workers. The experimental setup is shown in \autoref{fig:exp_setup}.
The relative positions of the objects and stations with respect to the robot were considered fixed. Moreover, the holographic instructions displayed by the Hololens 2 are also located in fixed relative poses with respect to the base of the manipulator, which is identified dynamically by the AR device at the beginning of the experiment. 
To meet the task requirements in terms of assembly precision, the autonomous behaviors of the robot were pre-planned. The target poses were read directly from the BT Robot Action nodes. The robot motions were executed with a quintic polynomial trajectory planner, in charge of sending suitable references to a low-level Cartesian position controller. Differently, collaborative actions were programmed in such a way that the manipulator could grasp the object autonomously and then switch to a Cartesian admittance controller. Thus, the worker could physically move the object or the cobot to accomplish the action while the cobot was lifting the object (completely or partially, depending on the object weight). Once the human had acknowledged the action completion, the robot could release the object and switch back to the Cartesian position controller. It is important to mention that not all the actions, but only a subset of the total job actions, were enabled as collaborative actions. This is due to the fact that the majority of the actions are simple and do not require two workers at the same time to be executed. However, when the action is complex and a multi-agent operation could be beneficial, a collaborative action execution is also considered. 
All experiments were performed at the Human-Robot Interfaces and Physical Interaction (HRII) laboratory, Istituto Italiano di Tecnologia (IIT), and the protocol was approved by the ethics committee of Azienda Sanitaria Locale (ASL) Genovese N.3 (Protocol IIT\_HRII\_ERGOLEAN 156/2020).

\subsubsection{Role Allocation Validation}\label{sssec:exp_hrc_ar}
\begin{figure}
    \centering
    \includegraphics[trim=0.8cm 0.5cm 0cm 2cm,clip,width=0.9\linewidth]{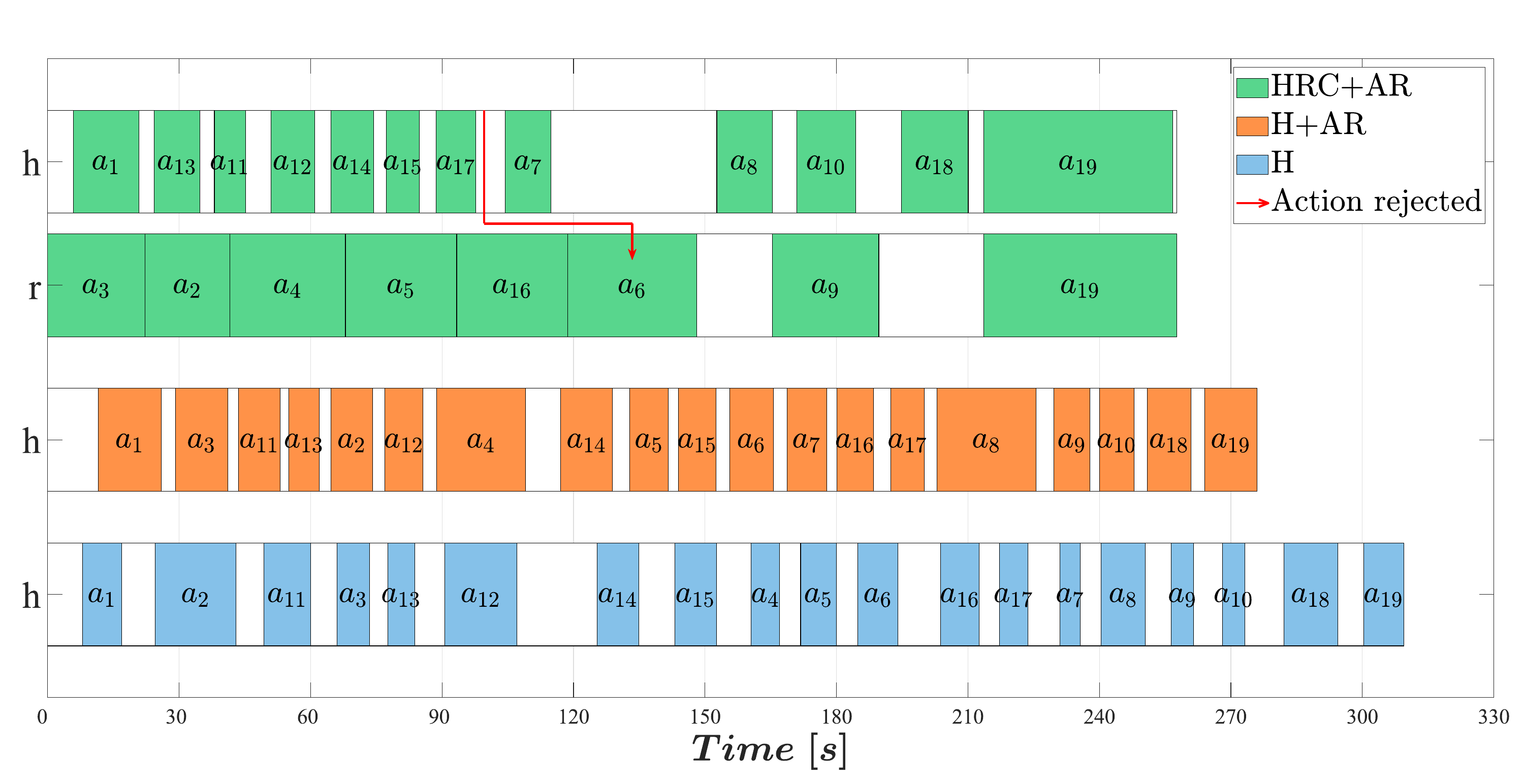}
    \caption{Detail of the Gantt charts related to the allocation of actions.}
    \label{fig:gantt_chart_real}
    \vspace{-2mm}
\end{figure}

In this experiment, the same subject executes the table assembly under different conditions: 
\begin{itemize}
    \item Autonomously, following a fixed plan, with written instruction on paper (denoted as \textit{H} which stands for human);
    \item Autonomously, with actions planned by the \textit{Collab-MT-BT} strategy, instructed by the AR interface (\textit{H+AR});
    \item In human-robot collaboration, with actions planned and assigned by the \textit{Collab-MT-BT} strategy, instructed by the AR interface (\textit{HRC+AR}).
\end{itemize}
The costs of each action associated with each worker, including the collaborative couples, are defined as the average time required to accomplish that specific action by the worker(s), similarly to the previous experiments. In particular, the worker and the robot executed each action 10 times, and the duration has been averaged and rounded to the nearest integer.
The costs and results of the allocations are illustrated in \autoref{table:allocation_results_real_exp}. For simplicity, when a worker was not enabled to execute an action, no cost is reported in the table (-). In the \textit{Worker allocated} column, $h\to r$ means that the action was allocated to the human ($h$) but was rejected and then is allocated to the robot ($r$). To compare the job execution in each condition, a detailed Gantt chart is plotted in \autoref{fig:gantt_chart_real}. It can be noticed that, overall, even if the single action duration result longer, the introduction of the AR interface promotes a faster job execution, by reducing the instruction-to-work reaction time. The increase in time in job execution is due to the introduction of the negotiation mechanism, in which the worker needs to acknowledge twice the execution of the action, before and after the action is achieved. In the (\textit{HRC+AR}) condition, moreover, the duties were split between two workers further reducing the total job time, even if the manipulator takes longer than the human to execute the actions and the worker rejected $a_6$ which was later assigned to the robot. 


\begin{table}[!t]
\begin{center}
\begin{tabular}{|c|c||c|c|c|c|}
\hline
\textbf{Sym} &\textbf{Action} & $c_{h}$ & $c_{r}$ & $c_{h\Join r}$ & \makecell{\textbf{Worker}\\\textbf{allocated}} \\ 
\hhline{------}
$a_1$ & Move J1 in S1 & 19 & 28 & - & \cellcolor{blue!50} $h$ \\[0.7mm]
$a_2$ & Insert L33\_1 in J1 & 22 & 24 & - & \cellcolor{green!50} $r$ \\[0.7mm]
$a_3$ & Move J2 in S2 & 19 & 27 & - & \cellcolor{green!50} $r$ \\[0.7mm]
$a_4$ & \makecell{Build Apron\_1: \\ insert J1+L33\_1 in J2} & 25 & 31 & - & \cellcolor{green!50} $r$ \\[0.7mm]
$a_5$ & \makecell{Lay down Apron\_1 \\ in S1-S2} & 20 & 30 & 41 & \cellcolor{green!50} $r$ \\[0.7mm]
$a_6$ & Insert G33\_1 in S1 & 20 & 34 & - & \cellcolor{green!50} $h\to r$ \\[0.7mm]
$a_7$ & Insert G33\_2 in S2 & 20 & 24 & - & \cellcolor{blue!50} $h$ \\[0.7mm]
$a_8$ & \makecell{Lay down Apron\_1 \\in S1-S2} & 26 & - & - & \cellcolor{blue!50} $h$ \\[0.7mm]
$a_9$ & Insert L50\_1 in S2 & 26 & 29 & - & \cellcolor{green!50} $r$ \\[0.7mm]
$a_{10}$ & Insert L50\_2 in S1 & 26 & 33 & - & \cellcolor{blue!50} $h$ \\[0.7mm]
$a_{11}$ & Move J3 in S3 & 19 & 28 & - & \cellcolor{blue!50} $h$ \\[0.7mm]
$a_{12}$ & Insert L33\_2 in J3 & 22 & 31 & - & \cellcolor{blue!50} $h$ \\[0.7mm]
$a_{13}$ & Move J4 in S4 & 19 & 30 & - & \cellcolor{blue!50} $h$ \\[0.7mm]
$a_{14}$ & \makecell{Build Apron\_2: \\ insert J3+L33\_2 in J4} & 25 & 31 & - & \cellcolor{blue!50} $h$ \\[0.7mm]
$a_{15}$ & \makecell{Lay down Apron\_2 \\ in S3-S4} & 25 & 31 & 35 & \cellcolor{blue!50} $h$ \\[0.7mm]
$a_{16}$ & Insert G33\_3 in S3 & 20 & 30 & - & \cellcolor{green!50} $r$ \\[0.7mm]
$a_{17}$ & Insert G33\_4 in S4 & 20 & 34 & - & \cellcolor{blue!50} $h$ \\[0.7mm]
$a_{18}$ & \makecell{Rotate 90° Apron\_2\\ in S3-S4} & 22 & - & - & \cellcolor{blue!50} $h$ \\[0.7mm]
$a_{19}$ & \makecell{Mount Apron\_2 \\ in S1-S2} & 25 & - & 15 & \cellcolor{orange!50} $h\Join r$ \\[0.7mm]
\hline
\end{tabular}
\end{center}
\caption{Allocation results of the actions of the first real world job. Costs represent the average execution time, expressed in seconds. Each color corresponds to a different allocated worker.}
\label{table:allocation_results_real_exp}
\vspace{-2mm}
\end{table}


\begin{table}[!t]
\begin{center}
\begin{tabular}{|c||c|c|c|c|c|}
\hline
\multicolumn{1}{|c||}{\multirow{2}{*}{\textbf{Sym}}} & \multicolumn{1}{|c|}{\multirow{2}{*}{$c^{init}_{\boldsymbol{h}}$}} & \multicolumn{1}{|c|}{\multirow{2}{*}{$c^{init}_{\boldsymbol{r}}$}} & \multicolumn{1}{|c|}{\multirow{2}{*}{$c^{init}_{\boldsymbol{h\;\Join\; r}}$}} & \multicolumn{2}{c|}{\textbf{Worker allocated}}\\[0.7mm]  \cline{5-6}
\multicolumn{1}{|c||}{} & & & & \textbf{HR Distance} & \textbf{Performance} \\[0.7mm]
\hhline{------}
$a_1$ & 40 & 20 & - & \cellcolor{green!50} $r$ & \cellcolor{blue!50} $h$\\[0.7mm]
$a_2$ & 40 & 20 & - & \cellcolor{blue!50} $h$ & \cellcolor{green!50} $r$\\[0.7mm]
$a_3$ & 40 & 20 & - & \cellcolor{blue!50} $h$ & \cellcolor{green!50} $r$\\[0.7mm]
$a_4$ & 40 & 20 & - & \cellcolor{green!50} $r$ & \cellcolor{green!50} $r$\\[0.7mm]
$a_5$ & 40 & 20 & 15 & \cellcolor{blue!50} $h$ & \cellcolor{blue!50} $h$\\[0.7mm]
$a_6$ & 40 & 20 & - & \cellcolor{blue!50} $h$ & \cellcolor{blue!50} $h$\\[0.7mm]
$a_7$ & 40 & 20 & - & \cellcolor{blue!50} $h$ & \cellcolor{green!50} $r$\\[0.7mm]
$a_8$ & 40 & - & - & \cellcolor{blue!50} $h$ & \cellcolor{blue!50} $h$\\[0.7mm]
$a_9$ & 40 & 20 & - & \cellcolor{blue!50} $h$ & \cellcolor{green!50} $r$\\[0.7mm]
$a_{10}$ & 40 & 20 & - & \cellcolor{green!50} $r$ & \cellcolor{blue!50} $h$\\[0.7mm]
$a_{11}$ & 40 & 20 & - & \cellcolor{blue!50} $h$ & \cellcolor{blue!50} $h$\\[0.7mm]
$a_{12}$ & 40 & 20 & - & \cellcolor{blue!50} $h$ & \cellcolor{blue!50} $h$\\[0.7mm]
$a_{13}$ & 40 & 20 & - & \cellcolor{green!50} $r$ & \cellcolor{blue!50} $h$\\[0.7mm]
$a_{14}$ & 40 & 20 & - & \cellcolor{green!50} $h\to r$ & \cellcolor{green!50} $h\to r$\\[0.7mm]
$a_{15}$ & 40 & 20 & 15 & \cellcolor{green!50} $r$ & \cellcolor{blue!50} $h$\\[0.7mm]
$a_{16}$ & 40 & 20 & - & \cellcolor{blue!50} $h$ & \cellcolor{green!50} $r$\\[0.7mm]
$a_{17}$ & 40 & 20 & - & \cellcolor{blue!50} $h$ & \cellcolor{blue!50} $h$\\[0.7mm]
$a_{18}$ & 40 & - & - & \cellcolor{blue!50} $h$ & \cellcolor{blue!50} $h$\\[0.7mm]
$a_{19}$ & 40 & - & 15 & \cellcolor{orange!50} $h\Join r$ & \cellcolor{orange!50} $h\Join r$\\[0.7mm]
\hline
\end{tabular}
\end{center}
\caption{Allocation results of the actions of the second real world job. Each color corresponds to a different allocated worker.}
\label{table:allocation_results_real_exp_distance}
\vspace{-2mm}
\end{table}

\subsubsection{Metrics Comparison}\label{sssec:exp_metrics}
So far, only action duration was used as cost function, but, as mentioned in \autoref{par:suitability_cost}, the method could feature other metrics as well. In this experiment, the effects on the allocation results due to the choice of the metric for the action cost are investigated. In particular, two different metrics are compared. The first, which is defined as \textit{Performance}, exploits the average action duration and is fixed during its execution (same costs as in \autoref{table:allocation_results_real_exp}). The second one, which changes dynamically, features the workers' positions in space (\textit{HR Distance}). In both cases, the condition \textit{HRC+AR} is employed.
It is common knowledge that the introduction of a robot in close proximity to a worker might induce stress on the operator. According to~\cite{arai2010assessment}, the amount of stress depends, among different factors, on the distance between the two workers and can influence the productivity of the task. In particular, the authors demonstrated that a higher strain on the human worker is required, both psychologically and physiologically, if the robot works at a distance smaller than 1 m.
In view of this, the action costs could be tuned proportionally with respect to the workers' distance in such a way that the largest possible, when the collaboration is not required, is maintained.  
One possible choice is the following:
\begin{equation} \label{eq:distance_costs}
    c_{ij} = \begin{cases}
    c^{init}_{ij}, \quad &\text{if } i=h \qquad \quad  \textit{(human)}, \\
    c^{init}_{ij} + \dfrac{\beta}{\lVert^h\boldsymbol{x}_{a_j}\rVert +\varepsilon}, \quad &\text{if } i=r \qquad \quad  \textit{(robot)}, \\
    c^{init}_{ij} + \gamma \lVert^h\boldsymbol{x}_{r}\rVert, &\text{if } i=h\Join r \quad \; \textit{(collaboration)},
    \end{cases}
\end{equation}
where $c^{init}_{ij}$ is the initial cost of the action $j$ for the worker $i$, $^h\boldsymbol{x}_{a_j}$ and $^h\boldsymbol{x}_{r}$ are the distances between the robot, the $j$-th action, and the human, respectively. $\beta$ and $\gamma$ are constant gains that allows $c_{ij}$ to be comparable with different $i$, i.e. in the same numeric range. To prevent numerical issues when $^h\boldsymbol{x}_{a_j}=0$, $\varepsilon>0$, small enough, is introduced. In this specific experiment, we selected $\beta = 20$, $\gamma = 35$ and $c^{init}_{ij}$ as in \autoref{table:allocation_results_real_exp_distance}. The idea is to penalize the robot's allocations to actions that are physically close to the current human position while favoring collaborative actions when the robot is in proximity to the human.
In general, this design and the proposed parameter values are not unique, and any other design with the same principles would fit this experiment.

\begin{figure}
    \centering
    \includegraphics[trim=0cm 0cm 0cm 0cm,clip,width=0.9\linewidth]{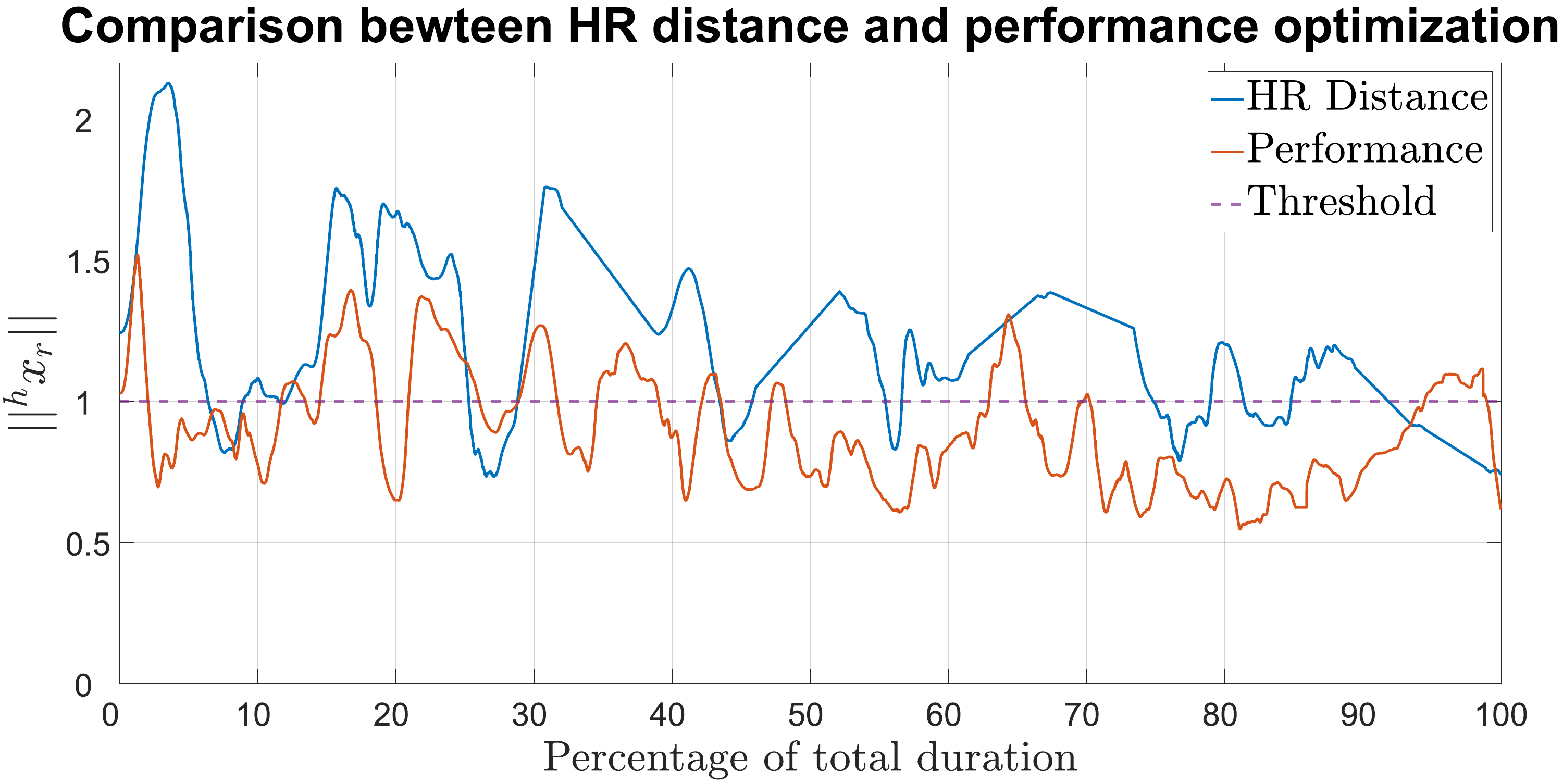}
    \caption{Human-robot distance during table assembly: comparison between \textit{Performance}, which uses predefined actions duration in \autoref{table:allocation_results_real_exp} as costs, and \textit{HR distance} costs computed with \eqref{eq:distance_costs}. The reference value of $1\;m$ represents the minimum distance between human and robot.}
    \label{fig:dist_perform_comp}
    \vspace{-2mm}
\end{figure}

In our setup, the system, thanks to the tracking capabilities of the Hololens 2, computes online the distance between the device and the virtual anchor, i.e., the base of the manipulator. Consequently, since the action locations are fixed, the distance between the human and the action and between the human and the robot end-effector can be computed. 
It's important to notice that this definition of the costs changes the allocation results but not the motion of the robot. 
The initial action-worker costs, $c^{init}_{ij}$, and the allocation results in both cases, are shown in \autoref{table:allocation_results_real_exp_distance}. Moreover, to evaluate the efficiency of the approach, the distance between the human and the robot, $\lVert^h\boldsymbol{x}_{r}\rVert$, is computed in both executions and it is plotted in \autoref{fig:dist_perform_comp} with the reference value ($1\;m$) \cite{arai2010assessment}.
On the x-axis, the percentage of total duration, is reported, since the two experiments have different duration ($169\;s$ the \textit{Performance}, $207\;s$ the \textit{HR Distance}). The last action, $a_{19}$, is not reported because is carried out by the workers in collaboration in both experiments (see \autoref{table:allocation_results_real_exp_distance}). As can be noticed in the plot, the distance between the workers is higher in the \textit{HR Distance} experiment than in the \textit{Performance} one. First, the average distance is above the threshold ($1.21\;m$) in the first case while less in the second one ($0.9\;m$). Moreover, the distance between the workers in the \textit{HR Distance} experiment is below the threshold only for $17.4\%$ of the total execution time, while the curve in the \textit{Performance} experiment has values below $1\;m$ of distance for almost half of the total job time ($49.2\%$). A video of the experiment with \textit{HR Distance} costs is available at \url{https://youtu.be/H3icROPSd8E} .

\subsubsection{User Testing Study}\label{sssec:exp_user_study}
\begin{figure}
    \centering
    \includegraphics[trim=0cm 0cm 0cm 0cm,clip,width=0.9\linewidth]{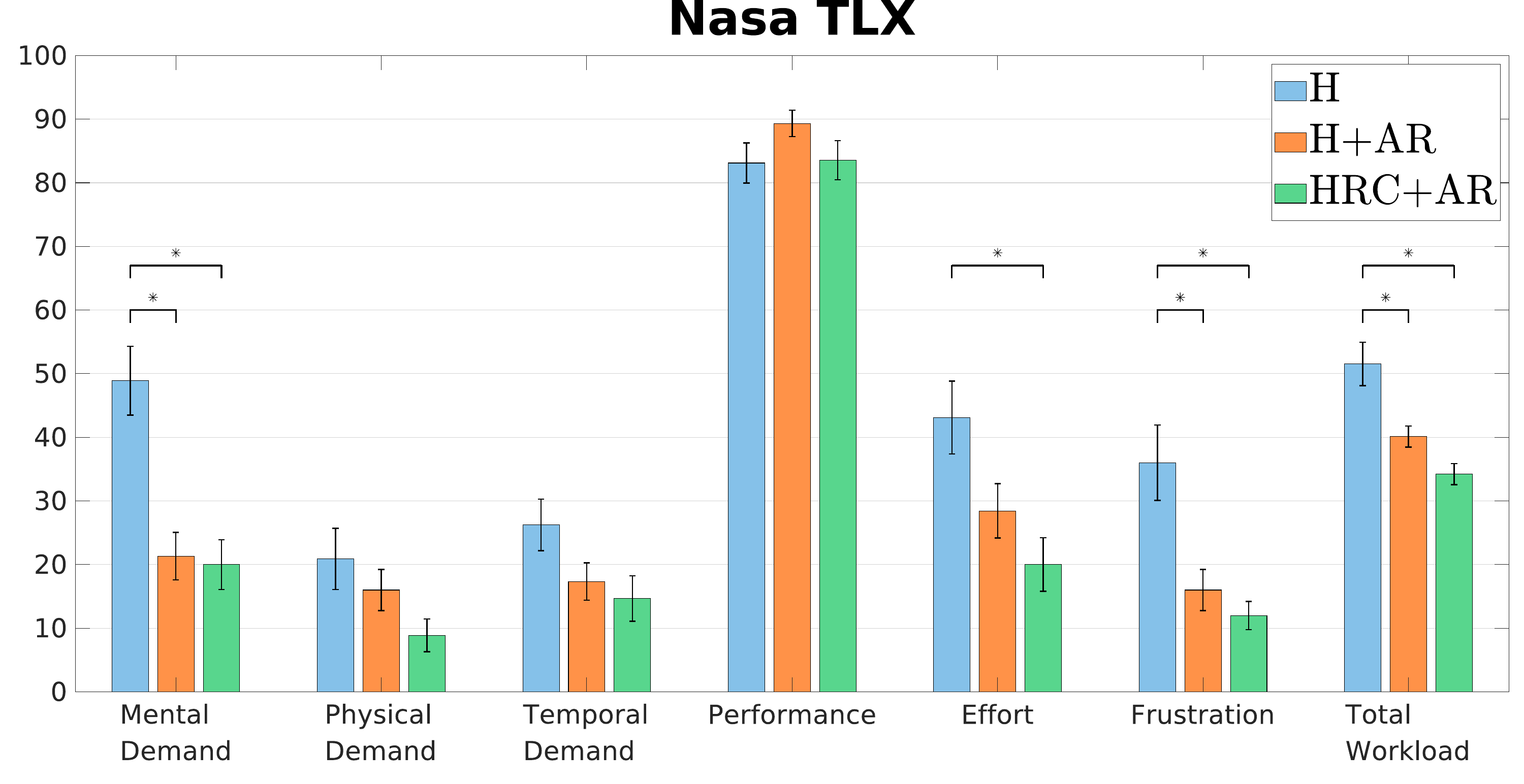}
    \caption{NASA TLX of the table assembly. The results are on a scale from 0 (very low) to 100 (very high). The color code is the same as in \autoref{fig:gantt_chart_real}.}
    \label{fig:nasa_tlx}
    \vspace{-2mm}
\end{figure}

To evaluate the usability and intuitiveness of the architecture, 25 healthy subjects, without any previous experience with the table assembly described in \autoref{sssec:exp_setup}, were asked to perform the latter under three different conditions, as detailed in \autoref{sssec:exp_hrc_ar}. The costs are the same as in \autoref{table:allocation_results_real_exp}. Before starting the experiment, each subject was provided with a short tutorial video which explains how to use the user interface (available at \url{https://youtu.be/EKDi1bEI2YQ}).
The order in which the subjects performed the experiment under the three conditions was randomly chosen and equally distributed in order to avoid affecting the results. To assess the workload imposed by the job, the 25 subjects, 15 males and 10 females ($27.8 \pm 2.3$ years old), were asked to fill out a NASA Task Load Index (NASA-TLX)~\cite{hart1988development}, at the end of each experiment (after each different condition) and a Likert scale-based custom questionnaire at the end of the whole experiment. 
NASA-TLX responses, under the three experimental conditions represented by the three different groups \textit{H}, \textit{H+AR}, and \textit{HRC+AR}, are scaled from 0 (very low) to 100 (very high), as illustrated in \autoref{fig:nasa_tlx}. The bars and error bars represent the mean of the scores selected by the subjects and the standard error of the mean, respectively. To statistically evaluate the results, first, an outlier removal was performed using the Grubbs test for outliers, which removes one outlier per iteration based on hypothesis testing. Then, to analyze the significance of the results (with $95\%$ confidence), the Analysis of Variance (ANOVA) test was selected. To do so, two assumptions have to be met: the one-sample Kolmogorov-Smirnov test and the Levene's test are executed for normality and equality of variance, respectively. The first condition was met for all data, while the second one was met only for the data related to Physical Demand, Temporal Demand, and Performance. Therefore, instead of the standard ANOVA test, a Welch ANOVA Test~\cite{welch1951comparison} was performed. From the tests, the results show a statistically significant difference in Mental Demand ($p = 0.00008$), Temporal Demand ($p = 0.0106$), Effort ($p = 0.0009$), Frustration ($p = 0.0021$), and Total Workload ($p = 0.00009$). Since the compared groups are 3 and the ANOVA test does not specify which group is significantly different from the others, the Bonferroni correction method was applied for the pairwise comparison of a multiple comparison test (ANOVA). Significant differences between groups, with confidence $95\%$, are represented by horizontal lines, pointing to the corresponding group, with a star on top. As can be noticed, the \textit{HRC+AR} experiment presents a reduction on all the scales but Performance, with respect to the human-only experiment (\textit{H}), which results in an overall significant reduction of the Total Workload.

\begin{figure}
    \centering
    \includegraphics[trim=0cm 0cm 0cm 0cm,clip,width=0.86\linewidth]{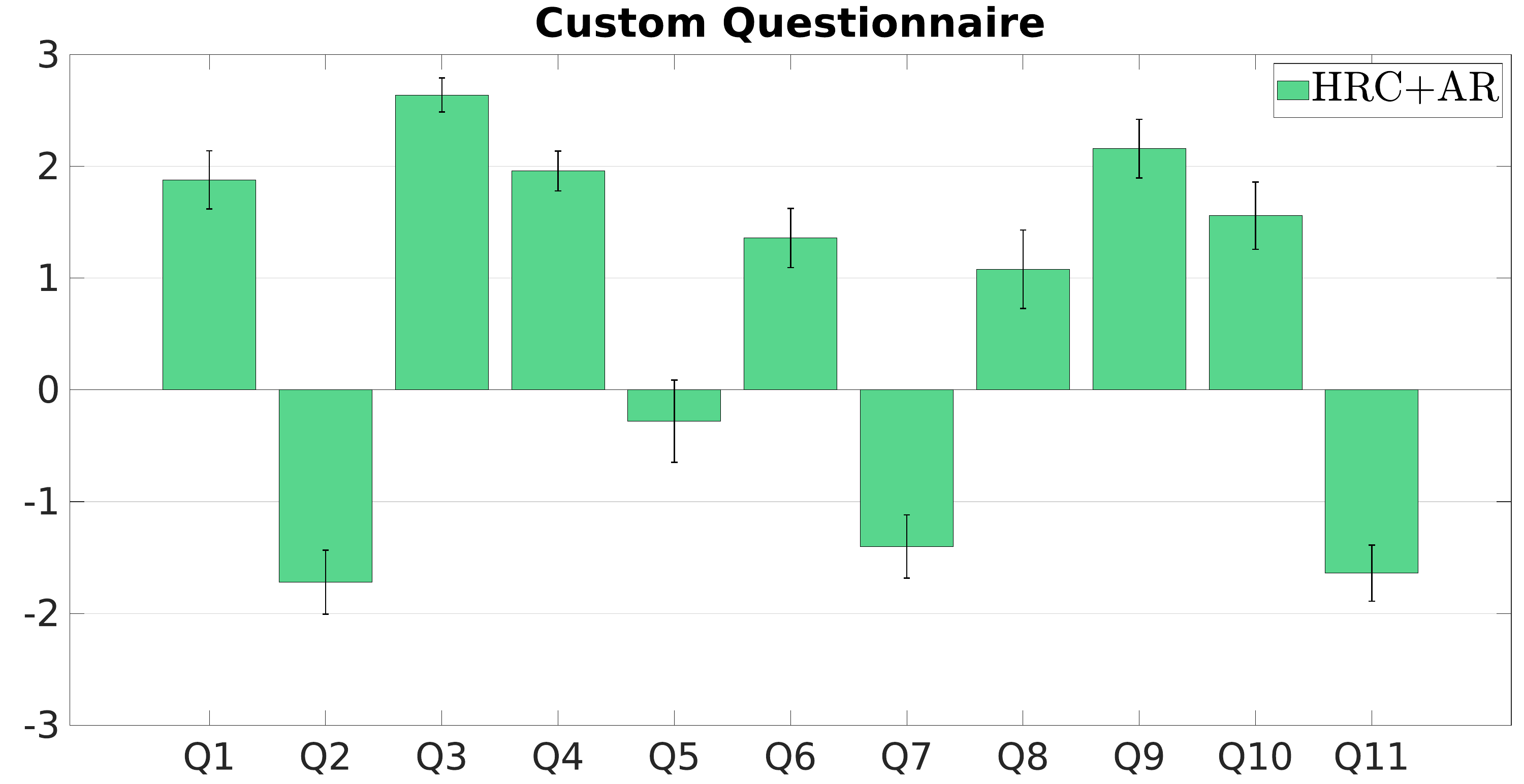}
    \caption{Custom Likert-scale questionnaire of the table assembly. The results are on a scale from -3 (strongly disagree) to 3 (strongly agree). The color code is the same as in \autoref{fig:gantt_chart_real}.}
    \label{fig:custom_quest}
    \vspace{-2mm}
\end{figure}

The custom questionnaire included 11 statements:
(Q1) It was easy to understand what action to take; (Q2) It was difficult to understand when I had to take action; (Q3) The robot was working while I was working; (Q4) It was easy to focus on action execution; (Q5) I had to wait for the system to inform me about the next action; (Q6) I felt I could delegate the actions to the robot; (Q7) During action execution, I experienced several periods of downtime; (Q8) The robot was able to adapt his action according to my decision; (Q9) I felt the robot was helping me to speed up action execution; (Q10) I appreciated executing some actions collaborating physically coupled with the robot; and (Q11) Overall, I would rather do the job alone.
Answers range uniformly from strongly disagreeing to strongly agree, with an assigned score of -3 and +3, respectively. \autoref{fig:custom_quest} shows the mean (bars) and the standard error on the mean (error bars) of the assigned scores selected by the subjects.
The participants agreed with the statements Q1, Q3, Q4, Q6, Q8, Q9, and Q10, while disagreeing with Q2, Q7, and Q11. Regarding Q5, instead, the subjects were neutral. Therefore, as for the NASA-TLX, the custom questionnaire validates the features of the proposed method: the architecture can maximize the parallelism of the actions between the workers, reducing their waiting times, and of the overall execution (Q3, Q7, Q9). Moreover, the approach allows the user to reject the actions and hence delegate them to the robot, while at the same time performing some actions in collaboration with the robot (Q6, Q8, Q10, Q11). The user interface, moreover, enables the rejection of the action and simplifies the worker's understanding of which action to take and how to perform it without distracting the user from the action execution (Q1, Q2, Q4).

%% file: sections/discussion.tex
In this manuscript, a unified architecture for dynamic role allocation and collaborative task planning in mixed human-robot teams was presented, suitable for flexible and fast reconfigurable industrial scenarios. The main advantages of the architecture are its high scalability across different industrial jobs (task-agnostic) and across motley teams in terms of size and agent heterogeneity.
The introduced architecture, thanks to the BT formalism, simplifies the modeling of human-robot collaborative processes, as it requires the same expertise required to design robotic planning problems. However, in our architecture, the BT is still manually designed, which means that profound knowledge of the job that has to be modeled is required. Moreover, only expert designers can achieve fast job deployment.

Through several simulations (\autoref{ssec:complexity}), the computational complexity was evaluated, and the results demonstrated a high suitability for industrial-sized jobs (up to $50$ actions) and medium-sized teams (up to $20$ workers) in case proper collaborations between workers are enabled (\textit{Collab-MT-BT}). In these conditions, the solver takes, overall, approx. $1\;s$ to compute the solution from the initialization of the BT to the allocation of the last action (\autoref{fig:complex_old_new_series}). That means, on average, less than $50\;ms$ to allocate a single action to a worker. In conditions of pure cooperation between workers (\textit{Coop-MT-BT}), similar performance is obtained with jobs sized up to $100$ actions, while no significant increase in terms of team size is reported (\autoref{fig:complex_agt}). With respect to the previous version of the allocation method (\textit{Coop-ST-BT})~\cite{fusaro2021integrated}, the approach (\textit{Collab-MT-BT}) properly exploits the job schedule in terms of action parallelization, as demonstrated in \autoref{ssec:simulations}. 
We acknowledge that this advantage is evident only in the presence of a job that allows a high degree of parallelism between actions.
Even in the presence of a unique centralized Role Allocator node, no relevant worsening in terms of computational time is recorded (see \autoref{fig:complex_old_new_series} and \autoref{fig:complex_old_new_parallel}). Overall, all the above mentioned methods, in the target jobs/team size that characterize industrial settings, outperform the gold standard in role allocation of assembly tasks represented by \textit{AND/OR graphs} \cite{johannsmeier2017hierarchical,lamon2019capability,darvish2018flexible,merlo2022dynamic} (\autoref{fig:complex_act}).

The different allocation results of the architecture evaluated with different action/worker suitability metrics (action duration and human-robot-actions distance, as in \autoref{fig:dist_perform_comp}) demonstrated the capability of the method in providing results that could optimize different criteria (total job duration and operator stress). Moreover, the proposed strategy, independently from the chosen cost metric and according to the job plan, has the potential to minimize workers' waiting times (see \autoref{fig:gantt_chart}).
While, on the one hand, the architecture includes collaborative assignments between agents, which are motivated by the recent development of collaborative technologies, the practical application of such solutions might be limited by two factors.
The first is related to the need to implement in the robot a newer execution mode, the collaborative one, which is often different with respect to the one the robot performs autonomously. For instance, in a collaborative transportation of a heavy box, the robot grasping point will likely differ if the box is also grasped by another agent.
The second might be the complexity of defining suitable metrics to measure collaborative performance. Often, it is not possible to exploit the single-agent capabilities to evaluate the teamed execution with costs defined in a model-based fashion. For this reason, in experiment III-C3, the collaborative cost is obtained with a different strategy with respect to the single-agent costs.

The User Testing study (\autoref{sssec:exp_user_study}) confirmed that the architecture has been perceived as beneficial by the subjects of the test, especially the allocation and planning mechanism (Q3, Q6, Q7, Q8, Q9, Q10). In particular, the subjects did not observe an increase in Temporal Demand (see also Q5 and Q6) or Frustration due to the presence of an external system that takes decisions in their place. Instead, Mental Demand results in a significant reduction. No significant differences relative to Physical Fatigue or Performance were observed.
Moreover, it can be noticed that the AR interface was perceived as an effective tool in giving instructions to human workers (Q1, Q4) and can contribute to the speed-up of the assembly process (\autoref{fig:gantt_chart_real}).
Finally, it is fundamental to distinguish between the proposed methodology and the current implementation with the available technologies. For example, we are aware that the current AR headset presents some physical limitations (e.g., comfort, weight, battery) that might prevent its use in industrial settings. However, we foresee technological advancements (hardware, software, and integration with IoT) that will make these devices common in the future.
Moreover, the promising results obtained demonstrate the potential of the technology in a laboratory setting and open the door to deeper studies in this direction.

%% file: sections/appendix.tex
\section{Algorithms Pseudocode}
\noindent For the sake of completeness, we include the pseudocode of the algorithms in \ref{ssec:hrc_model}.

\alglanguage{pseudocode}
\begin{algorithm}[!ht]
\small
\caption{Tick() function of the Agent Handler node.}
\label{algo:AgentHandler}
\begin{algorithmic}[1]
\Procedure{$\textproc{AgentHandler::tick}$}{$ $}
\State $getInput([W,\;A])$
\For{$[w,\;a]$ \textbf{in} $[W,\;A]$}
    \If {$w.type ==$ HUMAN $\lor$ $w.type ==$ COLLABORATIVE}
        \State \textbf{return} \emph{FAILURE}
    \ElsIf {$w.type ==$ ROBOT}
        \State \textbf{return} \emph{SUCCESS}
    \EndIf
\EndFor
\EndProcedure
\Statex
\end{algorithmic}
\vspace{-2mm}
\end{algorithm}

\alglanguage{pseudocode}
\begin{algorithm}[!ht]
\small
\caption{Tick() function of the Collaborative Handler node.}
\label{algo:CollaborativeHandler}
\begin{algorithmic}[1]
\Procedure{$\textproc{CollaborativeHandler::tick}$}{$ $}
\State $getInput([W,\;A])$
\For{$[w,\;a]$ \textbf{in} $[W,\;A]$}
    \If {$w.type ==$ COLLABORATIVE}
        \State \textbf{return} \emph{SUCCESS}
    \Else 
        \State \textbf{return} \emph{FAILURE}
    \EndIf
\EndFor
\EndProcedure
\Statex
\end{algorithmic}
\vspace{-2mm}
\end{algorithm}

\alglanguage{pseudocode}
\begin{algorithm}[!ht]
\small
\caption{Tick() function of the Action Completed node.}
\label{algo:ActionCompleted}
\begin{algorithmic}[1]
\Procedure{$\textproc{ActionCompleted::tick}$}{$ $}
\State \textproc{send}($a$ \textproc{is} $completed$)
\If {$response ==$ accepted}
    \State $w.available$ = \textbf{true}
    \State \textbf{return} \emph{SUCCESS}
\Else 
    \State \textbf{return} \emph{RUNNING}
\EndIf
\EndProcedure
\Statex
\end{algorithmic}
\vspace{-2mm}
\end{algorithm}